%% file: main.tex
\documentclass[11pt]{article} 
\usepackage{fullpage}
\usepackage{natbib}
\usepackage{float}
\usepackage[hyphens]{url}
\usepackage[hyphenbreaks]{breakurl}
\usepackage[colorlinks=true, allcolors=blue]{hyperref}
\usepackage{graphicx}
\usepackage{subfigure}

\usepackage{mathtools}
\input{math_commands}

\usepackage{tabularx,colortbl,xcolor,booktabs}
\allowdisplaybreaks
\usepackage{graphicx}
\usepackage{amsmath,amssymb,amsthm,amsfonts,enumitem,color}

\usepackage{mathtools}

\allowdisplaybreaks

\title{An Information Theoretic Approach to Interaction-Grounded Learning}

\author{
   Xiaoyan Hu\thanks{
	 The Chinese University of Hong Kong. Email: \texttt{xyhu21@cse.cuhk.edu.hk}}
   \and
   Farzan Farnia\thanks{
	 The Chinese University of Hong Kong. Email: \texttt{farnia@cse.cuhk.edu.hk}}
   \and
   Ho-fung Leung\thanks{Independent Researcher. Email: \texttt{ho-fung.leung@outlook.com}}
}

\begin{document}

\maketitle

\begin{abstract}
\input{0-abs}
\end{abstract}

\section{Introduction}
\label{sec:intro}
\input{1-intro}

\section{Related Work}
\label{sec:related_work}
\input{2-related}

\section{Preliminaries}
\label{sec:pre}
\input{3-preliminary}

\section{Variational Information-Based IGL}
\label{sec:vib-igl}
\input{4-VIB-IGL}

\section{The \texorpdfstring{$f$}{TEXT}-VI-IGL Algorithm}
\label{sec:alg}
\input{5-alg}

\section{Empirical Results}
\label{sec:emp}
\input{6-emp}

\section{Conclusion}
\label{sec:con}
\input{conclusion}

\bibliographystyle{apalike}
\bibliography{ref}

\newpage
\appendix
\input{Appendix}

\end{document}

%% file: math_commands.tex
\usepackage{amsmath,amssymb,amsthm,amsfonts,bm,bbm, nicefrac}
\usepackage{algorithm}
\usepackage{algorithmic}
\usepackage{xcolor}         
\definecolor{mydarkblue}{rgb}{0,0.08,0.45}
\definecolor{DarkBlue}{rgb}{0,0,0.55}
\definecolor{mygreen}{rgb}{0,0.7,0.7}

\DeclareMathAlphabet{\mathsfit}{\encodingdefault}{\sfdefault}{m}{sl}
\SetMathAlphabet{\mathsfit}{bold}{\encodingdefault}{\sfdefault}{bx}{n}

\def\gA{{\mathcal{A}}}

\def\gC{{\mathcal{C}}}
\def\gD{{\mathcal{D}}}

\def\gF{{\mathcal{F}}}
\def\gG{{\mathcal{G}}}

\def\gL{{\mathcal{L}}}

\def\gP{{\mathcal{P}}}

\def\gS{{\mathcal{S}}}
\def\gT{{\mathcal{T}}}

\def\gX{{\mathcal{X}}}
\def\gY{{\mathcal{Y}}}
\def\gZ{{\mathcal{Z}}}

\def\sP{{\mathbb{P}}}
\def\sQ{{\mathbb{Q}}}
\def\sR{{\mathbb{R}}}

\newcommand{\E}{\mathbb{E}}

\DeclareMathOperator*{\argmax}{arg\,max}
\DeclareMathOperator*{\argmin}{arg\,min}

\newenvironment{enumerate*}%
{\begin{enumerate}[leftmargin=*,topsep=0pt]%
		\setlength{\itemsep}{0pt}%
		\setlength{\parskip}{0pt}}%
	{\end{enumerate}}

\newcommand\indep{\protect\mathpalette{\protect\independenT}{\perp}}
\def\independenT#1#2{\mathrel{\rlap{$#1#2$}\mkern2mu{#1#2}}}

\theoremstyle{plain}
\newtheorem{theorem}{Theorem}[section]
\newtheorem{proposition}[theorem]{Proposition}

\theoremstyle{definition}
\newtheorem{definition}[theorem]{Definition}
\newtheorem{assumption}[theorem]{Assumption}
\theoremstyle{remark}
\newtheorem{remark}[theorem]{Remark}

\def\showcomments{1}
\ifnum\showcomments=1
\newcommand{\xy}[1]{[\textcolor{violet}{#1}]}

%% file: 0-abs.tex
Reinforcement learning (RL) problems where the learner attempts to infer an unobserved reward from some feedback variables have been studied in several recent papers. The setting of \textit{Interaction-Grounded Learning~(IGL)} is an example of such feedback-based RL tasks where the learner optimizes the return by inferring latent binary rewards from the interaction with the environment. In the IGL setting, a relevant assumption used in the RL literature is that the feedback variable $Y$ is conditionally independent of the context-action $(X,A)$ given the latent reward $R$. In this work, we propose \emph{Variational Information-based IGL (VI-IGL)} as an information-theoretic method to enforce the conditional independence assumption in the IGL-based RL problem. The VI-IGL framework learns a reward decoder using an information-based objective based on the conditional mutual information~(MI) between $(X,A)$ and $Y$. To estimate and optimize the information-based terms for the continuous random variables in the RL problem, VI-IGL leverages the variational representation of mutual information to obtain a min-max optimization problem. 
Also, we extend the VI-IGL framework to general $f$-Information measures leading to the generalized $f$-VI-IGL framework for the IGL-based RL problems. We present numerical results on several reinforcement learning settings indicating an improved performance compared to the existing IGL-based RL algorithm.

%% file: 1-intro.tex
In several applications of reinforcement learning (RL) algorithms, the involved agent lacks complete knowledge of the reward variable, e.g. in applications concerning brain-computer interface~(BCI)~\citep{1300799, 10.1007/978-3-319-46568-5_14} and recommender systems~\citep{maghakian2023personalized}. In such RL settings, the lack of an explicit reward could lead to a challenging learning task where the learner needs to infer the unseen reward from observed feedback variables. The additional inference task for the reward variable could significantly raise the computational and statistical complexity of the RL problem. Due to the great importance of addressing such RL problems with a misspecified reward variable, they have been exclusively studied in several recent papers~\citep{pmlr-v139-xie21e, xie2022interactiongrounded, maghakian2023personalized}.  

To handle the challenges posed by a misspecified reward variable, Xie~et~al.~\citeyearpar{pmlr-v139-xie21e, xie2022interactiongrounded} propose the \emph{Interaction-Grounded Learning~(IGL) framework}. According to the IGL framework, the agent observes a multidimensional \emph{context vector} based on which she takes an \emph{action}. Then, the environment generates a \emph{latent $0$-$1$ reward} and reveals a multidimensional \emph{feedback vector} to the agent. The agent aims to maximize the (unobserved) return by inferring rewards from the interaction, a sub-task which needs to be solved based on the assumptions on the relationship between reward and feedback variables. 

As a result, the key to addressing the IGL-based RL problem is a properly inferred \emph{reward decoder} $\psi\in\Psi$, which maps a context-action-feedback tuple $(X,A,Y)\in\gX\times\gA\times\gY$ to a prediction of the posterior probability on the latent reward $R$. Given such a reward decoder, the optimal policy can be obtained using standard contextual bandit algorithms~\citep{NIPS2007_4b04a686, Dud_k_2014}. However, such a reward decoder will be information-theoretically infeasible to learn without additional assumptions~\citep{xie2022interactiongrounded}. Consequently, the existing works on the IGL setting~\citep{pmlr-v139-xie21e,xie2022interactiongrounded} make relevant assumptions on the statistical relationship between the random variables of context $X$, action $A$, feedback $Y$, and latent reward $R$. In particular, a sensible assumption on the connection between $X,A,Y, R$ is the following conditional independence assumption proposed by \cite{pmlr-v139-xie21e}:

\begin{assumption}[Full conditional independence]
\label{as_fci}
For arbitrary $(X,A,R,Y)$ tuple where $R$ and $Y$ are generated based on the context-action pair $(X,A)$, the feedback $Y$ is conditionally independent of $X$ and $A$ given the latent reward $R$, i.e., $Y\indep X,A|R$.
\end{assumption} 

In the work of Xie~et~al.~\citeyearpar{pmlr-v139-xie21e}, a reward decoder $\psi:\gY\mapsto[0,1]$ takes the feedback $Y\in\gY$ as input and outputs a prediction of the posterior distribution $\sP(R=1|Y)$. Their proposed approach performs a joint training of the policy and the decoder by maximizing the difference in the decoded return between the learned policy and a ``bad'' policy that is known to have a low (true) return. They show that a properly inferred reward decoder can be learned statistically efficiently when: (i) the full conditional independence assumption~\ref{as_fci} \emph{strictly} holds, and (ii) the distributions $\sP(Y|R=0)$ and $\sP(Y|R=1)$ of the feedback variable $Y$ conditioned to the latent reward can be \emph{well separated}~\citep[Assumption 2]{pmlr-v139-xie21e}. However, these conditions are quite restricted in practice, where the observation of the feedback variable is often under significant noise levels, e.g. in the BCI application. In such noisy settings, Assumption~\ref{as_fci} may still hold under an independent noise from the discussed random variables or may not hold when the noise is correlated with the context or action variables. On the other hand, it can be much more difficult to distinguish between the feedback distributions conditioned to the latent reward. Consequently, the discussed IGL-based methods may no longer achieve optimal results under such noisy feedback conditions.

In this paper, we attempt to address the mentioned challenges in the IGL-based RL problem and propose \emph{Variational Information-based IGL (VI-IGL)} as an information-theoretic approach to IGL-based RL tasks. The proposed VI-IGL methodology is based on the properties of information measures that allow measuring the dependence among random variables. According to these properties, Assumption~\ref{as_fci} will hold, i.e., the feedback variable $Y$ is conditionally independent of the context-action $(X,A)$ given the latent reward $R$, \emph{if and only if} the conditional mutual information~(CMI) $I(Y;X,A|R)$ is zero. Therefore, we suggest an information bottleneck-based approach \citep{tishby2000information} and propose to learn a reward decoder via the following information-based objective value where $\beta>0$ is a tuning parameter and $R_\psi$ is the random decoded reward from $\psi$: 
\begin{equation}
\label{113}
    \argmin_{\psi\in\Psi}\{I(Y;X,A|R_\psi)-\beta\cdot I(X,A;R_\psi)\}
\end{equation}
Intuitively, minimizing the first term $I(Y;X,A|R_\psi)$ ensures that the solved reward decoder satisfies the full conditional independence assumption. In addition, the second term $I(R_\psi;X,A)$ serves as a regularization term ruling out naive reward decoders. 

Nevertheless, the objective function in~(\ref{113}) is challenging to optimize, since a first-order optimization of this objective requires estimating the value and derivatives of the MI for continuous random variables of the context $X$ and the feedback $Y$. To handle this challenge, we leverage the variational representation of MI~\citep{c52c83e0b02c4746a5ea29b5cd44fd00, 10.1109/TIT.2010.2068870} and cast Objective~(\ref{113}) as a min-max optimization problem that gradient-based algorithms can efficiently solve. Using the variational formulation of the information-based objective, we propose the Variational Information-based IGL (VI-IGL) minimax learning algorithm for solving the IGL-based RL problem. The VI-IGL method applies the standard gradient descent ascent algorithm to optimize the min-max optimization problem following the variational formulation of the problem. 

We numerically evaluate the proposed VI-IGL method on several RL tasks. Our empirical results suggest that VI-IGL can perform better than the baseline IGL RL algorithm in the presence of a noisy feedback variable. The main contributions of this paper can be summarized as:
\begin{enumerate}[leftmargin=*]
    \item We propose an information-theoretic approach to the IGL-based RL problem, which learns a reward decoder by minimizing an information-based objective function.
    \item To handle the challenges in estimating and optimizing ($f$-)MI for continuous random variables, we leverage the variational representation and formulate our objective as a min-max optimization problem, which can be solved via gradient-based optimization methods. We show that the optimal value can be sample-efficiently learned.
    \item We extend the proposed approach to $f$-Variational Information-based IGL~($f$-VI-IGL), leading to a family of algorithms to solve the IGL-based RL task. 
    \item We provide empirical results indicating that $f$-VI-IGL performs successfully compared to existing IGL-based RL algorithms.
\end{enumerate}

%% file: 2-related.tex
\textbf{Interaction-Grounded Learning (IGL).} The framework of IGL is proposed by~Xie~et~al.~\citeyearpar{pmlr-v139-xie21e} to tackle learning scenarios without explicit reward. At each round, the agent observes a multidimensional context, takes an action, and then the environment generates a latent 0-1 reward and outputs a multidimensional feedback. The agent aims to optimize the expected return by observing only the context-action-feedback tuple during the interaction. When the feedback is independent of both the context and the action given the latent reward~(full conditional independence), Xie~et~al.~show that the optimal policy can be sample-efficiently learned with additional assumptions. To relax the full conditional independence requirement, Xie~et~al.~\citeyearpar{xie2022interactiongrounded} introduce Action-Inclusive IGL, where the feedback can depend on both the latent reward and the action. They propose a contrastive learning objective and show that the latent reward can be decoded under a symmetry-breaking procedure. Recently, Maghakian~et~al.~\citeyearpar{maghakian2023personalized} apply the IGL paradigm with a multi-state latent reward to online recommender systems. Their proposed algorithm is able to learn personalized rewards and show empirical success.

\textbf{Information-Theoretic Reinforcement Learning Algorithms.} Reinforcement learning~(RL) is a well-established framework for agents' decision-making in an unknown environment~\citep{sutton2018reinforcement}. Several recent works focus on designing RL algorithms by exploiting the information-related structures in the learning setting. To perform exploration and sample-efficient learning, Russo and Van Roy~\citeyearpar{NIPS2014_301ad0e3} propose information-directed sampling~(IDS), where the agent takes actions that either with a small \emph{regret} or yield large \emph{information gain}, which is measured by the mutual information between the optimal action and the next observation. They show that IDS preserves numerous theoretical guarantees of Thompson sampling while offering strong performance in the face of more complex problems. 
In addition, information-theoretic approaches have been applied for \emph{skills discovery} in machine learning contexts. Gregor, Rezende, and Wierstra~\citeyearpar{gregor2016variational} introduce variational intrinsic control~(VIC), which discovers useful and diverse behaviors (i.e., \emph{options}) by maximizing the mutual information between the options and termination states. A setting that is close to our paper is using information-based methodology to learn reward functions in \emph{inverse reinforcement learning}~(IRL)~\citep{10.5555/645529.657801}. Levine, Popovi\'{c}, and Koltun~\citeyearpar{NIPS2011_c51ce410} propose to learn a cost function by maximizing the entropy between the corresponding optimal policy and human demonstrations. However, IGL is different from this setting, since it does not make any assumptions on the optimality of the observed behavior.

\textbf{Estimation of Mutual Information (MI).} Mutual information~(MI) is a fundamental information-theoretic quantity that measures ``the amount of information'' between random variables. However, estimating MI in continuous settings is statistically and computationally challenging \citep{pmlr-v38-gao15}. Building upon the well-known characterization of the MI as the Kullback-Leibler~(KL-) divergence~\citep{kullback1997information}, recent works propose to use the variational representation of MI for its estimation and more generally for $f$- divergences~\citep{10.1109/TIT.2010.2068870, pmlr-v80-belghazi18a, Song2020Understanding, 9053422}. An extra challenge in our paper is to estimate the mutual information conditioned to a latent variable.

%% file: 3-preliminary.tex
\subsection{Interaction-Grounded Learning (IGL)}

In the Interaction-Grounded Learning~(IGL) paradigm, at each round, a multidimensional \emph{context} $x\in\gX$ is drawn from a distribution $d_0$ and is revealed to the agent. Upon observing $x$, the agent takes action $a\in\gA$ from a finite action space. Let $\Delta_\gS$ denote the probability simplex on space $\gS$. Given the context-action pair $(x,a)$, the environment generates a \emph{latent and binary reward} $r\sim R(x,a)\in\Delta_{\{0,1\}}$ and returns a multidimensional \emph{feedback} $y\in\gY$ to the agent. It can be seen that IGL recovers a contextual bandit~(CB) problem~\citep{NIPS2007_4b04a686} if the reward is observed. Let $\pi\in\Pi:\gX\to\Delta_\gA$ denote any stochastic policy. The expected return of policy $\pi$ is given by $V(\pi):=\E_{x\sim d_0}\E_{a\sim\pi(\cdot|x)}[\mu(x,a)]$, where $\mu(x,a)$ is the expected (latent) reward of any context-action pair $(x,a)\in\gX\times\gA$. We consider batch mode learning, where the agent has access to a dataset $\{(x_k,a_k,y_k)\}_{k=1}^K$ collected by the behavior policy $\pi_\textup{b}:\gX\to\Delta_\gA$, where $x_k\sim d_0,a_k\sim\pi(\cdot|x_k)$, and $y_k$ is the stochastic feedback. The agent aims to learn the optimal policy, that is, $\pi^*:=\argmax_{\pi\in\Pi}V(\pi)$
while only observing the context-action-feedback tuple $(x,a,y)$ at each round of interaction.

\subsection{(\texorpdfstring{$f$}{TEXT}-)Conditional Mutual Information}
\label{sec_3.2}

The ($f$-)mutual information~(MI)~\citep{f3b5d8df-184a-3263-b0e6-61806ef005a0} is a standard measure of dependence between random variables in information theory. Formally, let $f:\sR_+\to\sR$ be a convex function satisfying $f(1)=0$. The $f$-MI~\citep{csiszar2022information} between a pair of random variables $Z_1$ and $Z_2$ is given by
\begin{equation}
\label{I_f-unconditional}
    I_f(Z_1;Z_2):=D_f(\sP_{Z_1Z_2}\|\sP_{Z_2}\otimes\sP_{Z_1}).
\end{equation}
In this definition, $D_f(\sP \| \sQ)$ denotes the $f$-divergence between distributions $\sP$ and $\sQ$ defined as
\begin{equation*}    D_f(\sP\|\sQ):=\E_{\sQ}\left[f\left(\frac{d\sP}{d\sQ}\right)\right]
\end{equation*}
Note that the standard KL-based conditional mutual information, which is denoted by $I(Z_1;Z_2)$, is given by $f(x)=x\log x$. Another popular $f$-divergence is Pearson-$\chi^2$~\citep{doi:10.1080/14786440009463897}, where $f(x)=(x-1)^2$. An important property of $f$-MI is that two random variables $Z_1,Z_2$ are statistically independent if and only $I_f(Z_1;Z_2)=0$, and hence dependence among between random variables can be measured via an $f$-mutual information. 

Furthermore, the $f$-conditional MI~\citep{csiszar2022information} between a pair of random variables $Z_1$ and $Z_2$ when $Z_3$ is observed can be defined as
\begin{equation}
\label{I_f}   I_f(Z_1;Z_2|Z_3):=D_f(\sP_{Z_1Z_2|Z_3}\|\sP_{Z_2|Z_3}\otimes\sP_{Z_1|Z_3}).
\end{equation}
Similarly, the standard KL-based conditional mutual information, denoted by $I(Z_1;Z_2|Z_3)$, is given by $f(x)=x\log x$. One useful property of the $f$-CMI is that, if $Z_1$ is conditionally independent of $Z_2$ given $Z_3$ then it holds that $I_f(Z_1;Z_2|Z_3)=0$.

%% file: 4-VIB-IGL.tex
In this section, we derive an information-theoretic formulation for the IGL-based RL problem. As discussed earlier, in information theory, a standard measure of the (conditional) dependence between random variables is (conditional) mutual information~(MI). Particularly, Assumption~\ref{as_fci}~(i.e., $Y\indep X,A|R$) is equivalent to that the conditional MI between the context-action $(X,A)$ and the feedback variable $Y$ is zero given the latent reward $R$, i.e., $I(Y;X,A|R)=0$. Here, we propose an information-theoretic objective function to learn a reward decoder $\psi\in\Psi:\gX\times\gA\times\gY\to[0,1]$
\begin{equation}
\label{obj}
    \argmin_{\psi\in\Psi}\{I(Y;X,A|R_\psi)-\beta\cdot I(X,A;R_\psi)\}
\end{equation}
where $\beta>0$ is a tunable hyperparameter. In the optimization of the above objective function, minimizing the first term $I(Y;X,A|R_\psi)$ guides the reward decoder to satisfy the conditional independence assumption. Furthermore, as the feedback variable is often under significant noise levels in practice, the second term $I(X,A;R_\psi)$ will play the role of a regularization term improving the robustness of the learned reward decoder against the noisy feedback.~(The detailed discussion can be found in Appendix~\ref{p_thm_vi-igl}.)

To handle the continuous random variables of the context $X$ and the feedback $Y$, we leverage the variational representation of the mutual information~\cite{10.1109/TIT.2010.2068870, pmlr-v80-belghazi18a} and reduce (\ref{obj}) to the following \emph{variational information-based IGL~(VI-IGL)} optimization problem.~ Here, we first present a min-max formulation for the above information-based optimization problem and a sample complexity bound for the resulting RL algorithm. Later in Section~\ref{sec:4.2}, we explain the steps in the proof.

\begin{theorem}[VI-IGL optimization problem]
\label{thm_vi-igl}

Objective~(\ref{obj})
\begin{equation*}
    \argmin_{\psi\in\Psi}\{I(Y;X,A|R_\psi)-\beta\cdot I(X,A;R_\psi)\}
\end{equation*}
is equivalent to the following optimization problem:
\begin{equation}
\label{opt_KL}
\begin{aligned}
    \argmin_{\psi\in\Psi}\; \gL(\psi):=\max_{G\in\gG}\min_{T\in\gT}\Big\{&\E_{\sP_{XAYR_\psi}}[G]-\E_{\sP_{Y|R_\psi}\otimes\sP_{XAR_\psi}}\left[e^{G}\right]\\
    &- \beta\cdot\left(\E_{\sP_{XAR_\psi}}[T]-\E_{\sP_{XA}\otimes\sP_{R_\psi}}\left[e^{T}\right]\right)\Big\}
\end{aligned}
\end{equation}
where
$G\in\gG:\gX\times\gA\times\gY\times\{0,1\}\to\sR$ and $T\in\gT:\gX\times\gA\times\{0,1\}\to\sR$ are two function classes.
\end{theorem}

\begin{algorithm}[htbp]
\begin{algorithmic}[1]
\caption{Variational Information-based IGL~(VI-IGL)}
\label{alg0}
\REQUIRE dataset $\gD_\textup{train}=\{(x_k,a_k,y_k)\}_{k=1}^K$, parameter $\beta>0$, reward decoder class $\Psi=\{\psi_\theta\}_{\theta\in\Theta}$, estimators $\gG=\{G_{\omega_1}\}_{\omega_1\in\Omega_1}$ and $\gT=\{T_{\omega_2}\}_{\omega_2\in\Omega_2}$ of $I(Y;X,A|R_\psi)$ and $I(X,A;R_\psi)$, respectively.
\FOR{epoch $k=1,2,\cdots,K$}
\STATE Sample a mini-batch $\gD_\textup{mini}\sim\gD_\textup{train}$.
\STATE Estimate the objective value $\widehat{\gL}(\psi)$ given by Equation~(\ref{opt_KL}).
\STATE Update the parameters
\begin{gather*}
    \omega_1\leftarrow\omega_1+\eta\cdot\nabla_{\omega_1}\widehat{\gL}(\psi)\\
    \omega_2\leftarrow\omega_2-\frac{\eta}{\beta}\cdot\nabla_{\omega_2}\widehat{\gL}(\psi)\\
    \theta\leftarrow\theta-\eta\cdot\nabla_\theta\widehat{\gL}(\psi)
\end{gather*}
where $\eta$ is the learning rate.
\ENDFOR
\STATE Train a policy $\pi$ via an offline contextual bandit oracle.
\STATE \textbf{Output:} Policy $\pi$.
\end{algorithmic}
\end{algorithm}

The inner level of the VI-IGL optimization problem minimizes over function class $\gT$ to estimate $I(X,A;R_\psi)$, the medium level maximizes over function class $\gG$ to estimate $I(Y;X,A|R_\psi)$, and the outer level minimizes over class $\Psi$ to find the appropriate reward decoder. 

Finally, as learning in IGL requires interaction with the environment, which can be expensive in practice, we provide theoretical guarantees for the VI-IGL optimization problem and show that the optimal objective value can be sample-efficiently learned. (The detailed proof can be found in Appendix~\ref{p_thm_sc}.)
\begin{theorem}[Sample complexity]
\label{thm_sc}
Consider a feedback-dependent reward decoder class $\Psi$ such that $\psi(y)\in[c,1-c]$ for any $\psi\in\Psi$ and $y\in\gY$, where $c\in(0,\frac{1}{2})$. Suppose the function classes $\gT$ and $\gG$ are bounded by $B\le\infty$. Then, for any $\delta\in(0,1]$, given a dataset $\gD=\{(x_k,a_k,y_k)\}_{k=1}^K$ collected by the behavior policy $\pi_\textup{b}:\gX\to\Delta_\gA$, there exists an algorithm such that the solved reward decoder $\widehat{\psi}$ from the optimization problem~(\ref{opt_KL}) satisfies that
\begin{equation}
\begin{aligned}
    \left|\gL(\widehat{\psi})-\gL^*\right|\le \max\{1,\beta\}\cdot O\Bigg(\frac{(1-c)^2}{c^2}\cdot \sqrt{\frac{\gC(\gY_\gP^\epsilon,B)}{K}\log\left(\frac{|\gY^\epsilon_{\gF}|d_{\Psi,\gT,\gG}}{\delta}\right)}\Bigg)
\end{aligned}
\end{equation}
where $\gL^*:=\min_{\psi\in\Psi}\gL(\psi)$ is the optimal value, $\gY^\epsilon_{\gF}\subset\gY$ is a $\epsilon$-covering of the feedback space $\gY$ equipped with (pseudo-)metric $\rho(y,y'):=\max_{F\in\gF}|F(y)-F(y')|$ where $\gF:=\Psi\cup\{e^{G(x,a,\cdot,r)}:(x,a,r)\in\gX\times\gA\times\{0,1\}\}_{G\in\gG}$, $\gC(\gY_\gP^\epsilon)$ is the capacity number defined in Equation~(\ref{eq_capa}) in the Appendix, and $d_{\Psi,\gT,\gG}$ is the statistical complexity of the joint function classes $\Psi$, $\gG$, and $\gT$, with the parameter $\epsilon=K^{-\nicefrac{1}{2}}$.
\end{theorem}

In practice, the functions $T,\, G$, and the reward decoder $\psi$ are often overparameterized deep neural networks which enable expressing complex functions. In application to deep neural networks, the covering number in the above sample complexity bound can be prohibitively large. We note that this issue in theoretically bounding the generalization error and sample complexity of deep learning algorithms is well-recognized in the supervised learning literature and is considered an open problem \cite{zhang2021understanding}. Similar to the supervised learning setting, we observed satisfactory numerical results achieved by the proposed VI-IGL-learned function, which highlights the role of gradient-based optimization in the success of the algorithm. Proving a sample complexity bound that takes the role of the gradient-based optimization algorithm into account will be an interesting future direction to our analysis.  


\subsection{Minimizing Conditional MI with Regularization}
\label{sec:4.1}

In this section, we present the detailed derivation of our information-theoretic objective~(\ref{obj}). Recall that we aim to learn a reward decoder $\psi\in\Psi:\gX\times\gA\times\gY\to[0,1]$ which minimizes the dependence measure $I(Y;X,A|R_\psi)$. Here, $R_\psi\sim\textup{Bernoulli}(\psi(X,A,Y))$ is the decoded 0-1 reward. On the other hand, note that the chain rule of MI results in the following identity
\begin{equation*}
\begin{aligned}
   I(Y;X,A|R_\psi)=I(Y;R_\psi|X,A)-I(Y;R_\psi)+I(Y;X,A)
\end{aligned}
\end{equation*}
As a result of the above information-theoretic identity, training to minimize only $I(Y;X,A|R_\psi)$ may result in a context-action-dependent reward decoder $\psi:\gX\times\gA\to[0,1]$, i.e., $I(Y;R_\psi|X,A)=0$, that ``over-fits'' to the feedback $Y$ to maximize $I(Y;R_\psi)$, and hence may underperform under a noisy feedback variable. To address this issue, we propose the regularized information-based IGL objective~(\ref{obj}) where $\beta>0$ is a tunable parameter:
\begin{equation*}
    \argmin_{\psi\in\Psi}\{I(Y;X,A|R_\psi)-\beta\cdot I(X,A;R_\psi)\}\tag{4}
\end{equation*}
To gain intuition on why Objective~(\ref{obj}) can be robust against noisy feedback, note that 
\begin{equation*}
    I(X,A;R_\psi)=H(R_\psi)-H(R_\psi|X,A)
\end{equation*}
where $H$ is the Shannon entropy. Thus, Objective~(\ref{obj}) encourages the reward decoder to remain unchanged to the context-action $(X,A)$ to minimize $H(R_\psi|X,A)$. Hence, the noises present in the feedback variable $Y$ cannot significantly affect the accuracy of the optimized reward decoder. 
On the other hand, we can show that in the noiseless setting, including the regularization term does not (greatly) affect the quality of the optimized reward decoder with a proper selection of $\beta$.~(The detailed proof can be found in Appendix~\ref{p_thm_4}.)

\begin{theorem}[Regularization (almost) ensures conditional independence]
\label{thm_4}
    Assume the reward decoder class $\Psi$ admits realizability assumption, i.e., there exists $\tilde{\psi}\in\Psi$ such that $I(Y;X,A|R_{\tilde{\psi}})=0$. Then, under Assumption~\ref{as_fci}, any reward decoder optimizing Objective~(\ref{obj}) satisfies that
    \begin{equation}
    \label{eq_thm4}
        I(Y;X,A|R_\psi)\le\beta\cdot(\log2-I(Y;R))
    \end{equation}
    where $R$ is the true latent binary reward and $I(Y;R)\le\log2$. Particularly, when $\Psi$ is feedback-dependent, a reward decoder $\psi:\gY\to[0,1]$ attains the minimum if and only if $I(Y;X,A|R_\psi)=0$.
\end{theorem}

In other words, for a feedback-dependent reward decoder class, the optimized reward decoder is guaranteed to satisfy the conditional independence assumption \emph{regardless of} the selection of $\beta$. For reward decoder class that also depends on the context-action, the learned reward decoder violates Assumption~\ref{as_fci} by at most (a multiplicative of) $\beta$. 

As demonstrated by our numerical results in Section \ref{sec: 6.2}, introducing this regularizer not only helps to handle a noisy feedback variable, but also results in a more consistent algorithm performance under lower noise levels. 

\subsection{Leveraging Variational Representation to Solve Information-based Objective}
\label{sec:4.2}

While the previous sub-section introduces an information-theoretic objective to address the IGL-based RL problem, optimizing (\ref{obj}) in complex environments can be highly challenging. The primary challenge to solve (\ref{obj}) is that it requires estimating MI among continuous random variables of the context $X$ and the feedback $Y$, which is widely recognized as a statistically and computationally difficult problem~\citep{10.1162/089976603321780272}. To derive a tractable optimization problem, we utilize the variational representation of the KL-divergence, which reduces the evaluation and estimation of MI to an optimization task.

\begin{proposition}[Donsker-Varadhan representation~\cite{c52c83e0b02c4746a5ea29b5cd44fd00}]
\label{prop_vrkl}
Let $\sP,\sQ\in\Delta_\gS$ be two probability distributions on space $\gS$. Then,
\begin{equation*}
\begin{aligned}
    D_\textup{KL}(\sP\|\sQ)=\sup_{T\in\gT}\left\{\E_{s\sim\sP}[T(s)]-\E_{s\sim\sQ}\left[e^{T(s)}\right]\right\}
\end{aligned}
\end{equation*}
where the supremum is taken over all functions T such that the two expectations are ﬁnite.
\end{proposition}

Recall that the MI between random variables $Z_1\in\gZ_1$ and $Z_2\in\gZ_2$ is the KL-divergence between their joint distribution $\sP_{Z_1Z_2}$ and the product of their marginal distributions $\sP_{Z_1}\otimes\sP_{Z_2}$, i.e., $I(Z_1;Z_2)=D_\textup{KL}(\sP_{Z_1Z_2}\|\sP_{Z_1}\otimes\sP_{Z_2})$. Proposition~\ref{prop_vrkl} enables us to estimate $I(Z_1;Z_2)$ through optimizing over a class of function $T:\gZ_1\times\gZ_2\to\sR$. Therefore, directly applying Proposition~\ref{prop_vrkl} to Objective~(\ref{obj}) results in the VI-IGL optimization problem in Theorem~\ref{thm_vi-igl}.
\begin{equation*}
\begin{aligned}
    \argmin_{\psi\in\Psi}\max_{G\in\gG}\min_{T\in\gT}\Big\{&\E_{\sP_{XAYR_\psi}}[G]-\E_{\sP_{Y|R_\psi}\otimes\sP_{XAR_\psi}}\left[e^{G}\right]- \beta\cdot\left(\E_{\sP_{XAR_\psi}}[T]-\E_{\sP_{XA}\otimes\sP_{R_\psi}}\left[e^{T}\right]\right)\Big\}
\end{aligned}
\end{equation*}

%% file: 5-alg.tex
\subsection{The Extended \texorpdfstring{$f$}{TEXT}-Variational Information-based IGL}
In this section, we first propose an extended version of the information-based objective in (\ref{obj}) and the VI-IGL optimization problem~(\ref{opt_KL}). Recall that $f$-mutual information defined in Equation~(\ref{I_f}) generalizes the standard KL-divergence-based MI to a general $f$-divergence-based MI. Therefore, we can extend the standard MI-based IGL objective~(\ref{obj}) to the following $ f$-MI-based IGL objective:
\begin{equation}
\label{obj-f}
    \psi^*:=\argmin_{\psi\in\Psi}\{I_{f_1}(Y;X,A|R_\psi)-\beta\cdot I_{f_2}(X,A;R_\psi)\}
\end{equation}
where $f_1$ and $f_2$ are two $f$-divergences. Note that Objective~(\ref{obj}) is a special case of the above formulation by selecting $f_1(x)=f_2(x)=x\log x$ to obtain the standard KL-based mutual information. 
Similar to the VI-IGL problem formulation, to derive a tractable optimization problem corresponding to the above task, we adopt the variational representation of $f$-divergences~(Proposition~\ref{prop_varrep} in Appendix~\ref{sec_var}). We propose the following min-max optimization problem to solve Objective~(\ref{obj-f}).

\begin{theorem}[$f$-VI-IGL optimization problem]
Let $f_1$ and $f_2$ be functions satisfying the requirements in Proposition~\ref{prop_varrep} and we denote by $f^*_1$ and $f_2^*$ their Fenchel conjugate, respectively. Objective~(\ref{obj-f}) is equivalent to the following min-max optimization problem
\begin{equation}
\label{opt_obj}
\begin{aligned}
    \min_{\psi\in\Psi}\max_{G\in\gG}&\min_{T\in\gT}\Big\{\E_{\sP_{XAYR_\psi}}[G]-\E_{\sP_{Y|R_\psi}\otimes\sP_{XAR_\psi}}[f_1^*(G)]- \beta\cdot(\E_{\sP_{XAR_\psi}}[T])-\E_{\sP_{XA}\otimes\sP_{R_\psi}}[f_2^*(T)])\Big\}
\end{aligned}
\end{equation}
where $G\in\gG:\gX\times\gA\times\gY\times\{0,1\}\to\sR$ and $T\in\gT:\gX\times\gA\times\{0,1\}\to\sR$.
\end{theorem}
Similarly, we derive the sample complexity for the above optimization problem in Theorem~\ref{thm_sc_f} in the Appendix.

\begin{algorithm}[t]
\begin{algorithmic}[1]
\caption{$f$-Variational Information-based IGL~($f$-VI-IGL)}
\label{alg}
\REQUIRE dataset $\gD_\textup{train}=\{(x_k,a_k,y_k)\}_{k=1}^K$, parameter $\beta>0$, reward decoder class $\Psi=\{\psi_\theta\}_{\theta\in\Theta}$, estimators $\gG=\{G_{\omega_1}\}_{\omega_1\in\Omega_1}$ and $\gT=\{T_{\omega_2}\}_{\omega_2\in\Omega_2}$ of $I_{f_1}(Y;X,A|R_\psi)$ and $I_{f_2}(X,A;R_\psi)$, respectively.
\FOR{epoch $k=1,2,\cdots,K$}
\STATE Sample a mini-batch $\gD_\textup{mini}\sim\gD_\textup{train}$.
\STATE Construct datasets with distributions $\sP_Y\otimes\sP_{R_{\psi_\theta}}$ and $\sP_{Y|R_{\psi_\theta}}\otimes\sP_{XAR_{\psi_\theta}}$ using $\gD_\textup{mini}$~(See the algorithm description).
\STATE Estimate the $f$-MI terms
\begin{equation*}
    \widehat{I}_{f_1}(X,A;R_{\psi_\theta})\leftarrow\E_{\sP_{XAR_{\psi_\theta}}}[T]-\E_{\sP_{XA}\otimes\sP_{R_{\psi_\theta}}}[f_1^*(T)]
\end{equation*}
\begin{equation*}
\begin{aligned}
    \widehat{I}_{f_2}(Y;X,A|R_{\psi_\theta})\leftarrow&\E_{\sP_{XAYR_{\psi_\theta}}}[G]-\E_{\sP_{Y|R_{\psi_\theta}}\otimes\sP_{XAR_{\psi_\theta}}}[f_2^*(G)]
\end{aligned}
\end{equation*}
\STATE Update the parameters
\begin{gather*}
    \omega_1\leftarrow\omega_1+\eta\cdot\nabla_{\omega_1}\left\{\widehat{I}_{f_1}(X,A;R_\psi)\right\}\\
    \omega_2\leftarrow\omega_2+\eta\cdot\nabla_{\omega_2}\left\{\widehat{I}_{f_2}(X,A;R_\psi)\right\}\\
    \theta\leftarrow\theta-\eta\cdot\nabla_\theta\left\{\widehat{I}_{f_1}(Y;X,A|R_{\psi_\theta})-\beta\cdot\widehat{I}_{f_2}(X,A;R_{\psi_\theta})\right\}
\end{gather*}
where $\eta$ is the learning rate.
\ENDFOR
\STATE Select between $\psi_\theta$ and its opposite counterpart $1-\psi_\theta$ based on their decoded returns of $\pi_b$.
\STATE Train a policy $\pi$ via an offline contextual bandit oracle.
\STATE \textbf{Output:} Policy $\pi$.
\end{algorithmic}
\end{algorithm}

\subsection{Algorithm Description}
\label{sec: 5.2}
Here, we present $f$-VI-IGL Algorithm~\ref{alg} as an optimization method to solve the $f$-VI-IGL optimization problem~(\ref{opt_obj}) for continuous random variables of the context $X$ and the feedback $Y$. The algorithm optimizes over three function classes $\gG,\gT$, and $\Psi$. Specifically, function class $\Psi=\{\psi_\theta\}_{\theta\in\Theta}$ consists of the reward decoders parameterized by $\theta\in\Theta$. Function class $\gG=\{G_{\omega_1}\}$ parameterized by $\omega_1\in\Omega_1$ is the estimator of $f_1$-MI $I_{f_1}(Y;X,A|R_{\psi_\theta})$. In addition, function class $\gT=\{T_{\omega_2}\}_{\omega_2\in\Omega_2}$ parameterized by $\omega_2\in\Omega_2$ is the estimator of $f_2$-MI $I_{f_2}(X,A;R_{\psi_\theta})$. We focus on learning in the batch mode, where the algorithm has access to an offline dataset $\gD_\textup{train}=\{(x_t,a_t,y_t)\}_{t=1}^T$ consisting of the context-action-feedback tuples, which is collected by the behavior policy $\pi_b$ interacting with the environment. 

At each epoch, $f$-VI-IGL first uses a mini-batch of data to estimate the value of Objective~(\ref{opt_obj})~(Lines 2-4). One difficulty is that estimating $I_{f_1}(Y;X,A|R_{\psi_\theta})$ requires sampling $(x,a,y)\sim\sP_{R_{\psi_\theta}}\otimes\sP_{Y|R_{\psi_\theta}}\otimes\sP_{XA|R_{\psi_\theta}}$, where $\sP_{Y|R_{\psi_\theta}}$ and $\sP_{XA|R_{\psi_\theta}}$ can be intractable for continuous random variables of the context $X$ and the feedback $Y$. To address the problem, we first augment each sample $(x_t,a_t,y_t)$ $N$ times to obtain $\{(x_t,a_t,y_t,r_t^i)\}_{i=1}^N$, where $r_t^i\sim\textup{Bernoulli}(\psi_\theta(x_t,a_t,y_t))$ and $N$ is a small positive integer~(e.g. 5 in our experiments). To sample, e.g., the feedback $y\sim\sP_{Y|R_{\psi_\theta}=1}$, we randomly sample a data point from $\{(x_t,a_t,y_t,r_t^j):j\in[N],r_t^j=1\}_{t=1}^T$, i.e., the ``augmented'' data points whose random decoded reward is 1. Given the estimated objective value, we alternatively update the parameters for the $f$-MI estimators and the reward decoder~(Line 5). At the end of the training, we use the learned reward decoder $\psi_\theta$ to train a policy via an offline contextual bandit oracle~\citep{NIPS2007_4b04a686, dudik2011efficient}. However, note that in Objective~(\ref{obj-f}), both the optimal reward decoder $\phi^*$ and its opposite counterpart $1-\phi^*$ may attain the minimum simultaneously~(while only one of them aligns is consistent with the true latent reward). Hence, we use the data-driven collector~\citep{pmlr-v139-xie21e} and select the reward decoder~(between the learned reward decoder $\psi_\theta$ and its opposite counterpart $1-\psi_\theta$) that gives a decoded return of $\pi_b$ lower than 0.5.\footnote{Following the previous works~\citep{pmlr-v139-xie21e, xie2022interactiongrounded}, we assume the behavior policy has a low (true) return.}

%% file: 6-emp.tex
In this section, we numerically evaluate the $f$-VI-IGL algorithm on the number-guessing task~\citep{pmlr-v139-xie21e} with noisy feedback, the details of which are described in the following.

\textbf{Number-guessing task with noisy feedback.} In the standard setting, a \emph{random} image $x_t$~(context), whose corresponding number is denoted by $l_{x_t}\in\{0,1,\cdots,9\}$, is drawn from the MNIST dataset~\citep{Lecun1998} at the beginning of each round $t$. Upon observing $x_t$, the learner selects $a_t\in\{0,1,\cdots,9\}$ as the predicted number of $x_t$~(action). The latent binary reward $r_t=\mathbbm{1}[a_t=l_{x_t}]$ is the correctness of the prediction label. Then, a \emph{random} image of digit $r_t\in\{0,1\}$ is revealed to the learner~(feedback). In many real-world scenarios, the observation of the feedback variable is often under significant noise level, e.g., in the BCI application. To simulate these cases, we consider four types of noisy feedback. Specifically, with a small probability, the feedback is replaced with: 1) \emph{independent noises}~(\texttt{I}): a random image of letter ``t''~(\emph{True}) when the guess is correct or a random image of letter ``f''~(\emph{False}) when the guess is wrong, which is sampled from the EMNIST Letter dataset~\citep{cohen2017emnist}, 2) \emph{action-inclusive noises}~(\texttt{A}): a random image of digit $(a_t+6\cdot r_t-3)\textup{ mod }10$, 3) \emph{context-inclusive noises}~(\texttt{C}): a random image of digit $(l_{x_t}+6\cdot r_t-3)\textup{ mod }10$, 4) \emph{context-action-inclusive noises}~(\texttt{C-A}): a random image of digit $(l_{x_t}+a_t+6\cdot r_t-3)\textup{ mod }10$. An example is given in Table~\ref{exp_set}. Note that the full conditional independence assumption does not strictly hold as the feedback is also affected by the context-action pair~(except for the independent noises).

\begin{table}[htbp]
\begin{center}
\begin{tabular}{ |l||c|c|c|c| }
\hline
$X=$\begin{minipage}{.05\textwidth}
      \includegraphics[width=8mm, height=8mm]{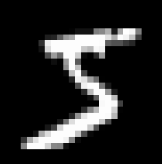}
    \end{minipage} &
\texttt{I} & \texttt{A} & \texttt{C} & \texttt{C-A} \\
\hline
\hline
Noisy $Y$ ($A=5$)   &  \begin{minipage}{.05\textwidth}
      \includegraphics[width=8mm, height=8mm]{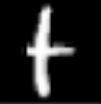}
    \end{minipage} & \begin{minipage}{.05\textwidth}
      \includegraphics[width=8mm, height=8mm]{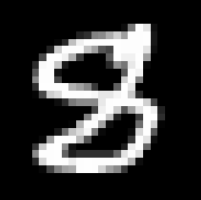}
    \end{minipage} & \begin{minipage}{.05\textwidth}
      \includegraphics[width=8mm, height=8mm]{8.png}
    \end{minipage} & \begin{minipage}{.05\textwidth}
      \includegraphics[width=8mm, height=8mm]{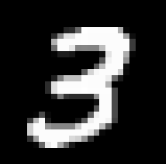}
    \end{minipage} \\
\hline
\hline
Noisy $Y$ ($A=6$)   &  \begin{minipage}{.05\textwidth}
      \includegraphics[width=8mm, height=8mm]{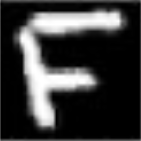}
    \end{minipage} & \begin{minipage}{.05\textwidth}
      \includegraphics[width=8mm, height=8mm]{3.png}
    \end{minipage} & \begin{minipage}{.05\textwidth}
      \includegraphics[width=8mm, height=8mm]{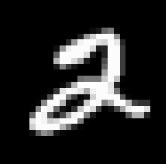}
    \end{minipage} & \begin{minipage}{.05\textwidth}
      \includegraphics[width=8mm, height=8mm]{8.png}
    \end{minipage} \\
\hline
\end{tabular}
\caption{An example of the noisy feedback: The context is a random image of digit ``5'', i.e., $l_x=5$. The rows show the noisy feedback when the guess is digit ``5''~(correct) and ``6''~(wrong), respectively. The columns show the cases for each type of noise~(\texttt{I}: independent, \texttt{A}: action-inclusive, \texttt{C}: context-inclusive, \texttt{C-A}: context-action-inclusive).}
\label{exp_set}
\end{center}
\end{table}

\textbf{Data collection.} We focus on the batch learning mode, where a training dataset $\gD_\textup{train}=\{(x_k,a_k,y_k)\}_{k=1}^K$ is collected by the uniform behavior policy using the \emph{training set}. In all the experiments, the training dataset contains $60,000$ samples, i.e., $K=60,000$. The output (linear) policy is evaluated on a test dataset $\gD_\textup{test}$ containing $10,000$ samples of context, which is randomly collected from the \emph{test set}. Additional experimental details are provided in Appendix~\ref{App_A}.

\subsection{Robustness to Noises}
\label{sec: emp_noises}

In this section, we show that VI-IGL optimizing the standard MI-based Objective~(\ref{obj}) is more robust to the noisy feedback than the previous IGL-based \textsf{E2G} algorithm~\citep{pmlr-v139-xie21e}. We compare the accuracy of the output policy under different noise levels~($10\%,20\%,30\%$), and the results are summarized in Figure~\ref{fig:1}.~(The detailed data can be found in Appendix~\ref{sec:F.1}) In the noiseless setting, VI-IGL achieves a comparable performance~($(81.6\pm7.9)\%$) to \textsf{E2G}~($(82.2\pm4.3)\%$). However, VI-IGL~(blue lines) significantly outperforms \textsf{E2G}~(orange lines) \emph{in all noisy settings and across all noise levels}. 

\begin{figure}[h]
\centering
    \subfigure[Independent]{\includegraphics[width=0.35\textwidth]{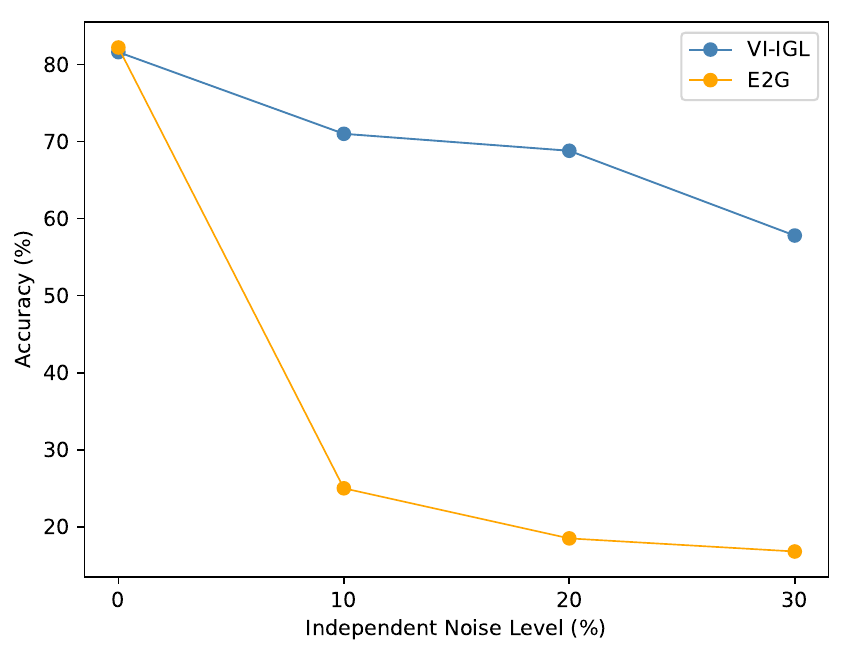}} 
    \subfigure[Action-inclusive]{\includegraphics[width=0.35\textwidth]{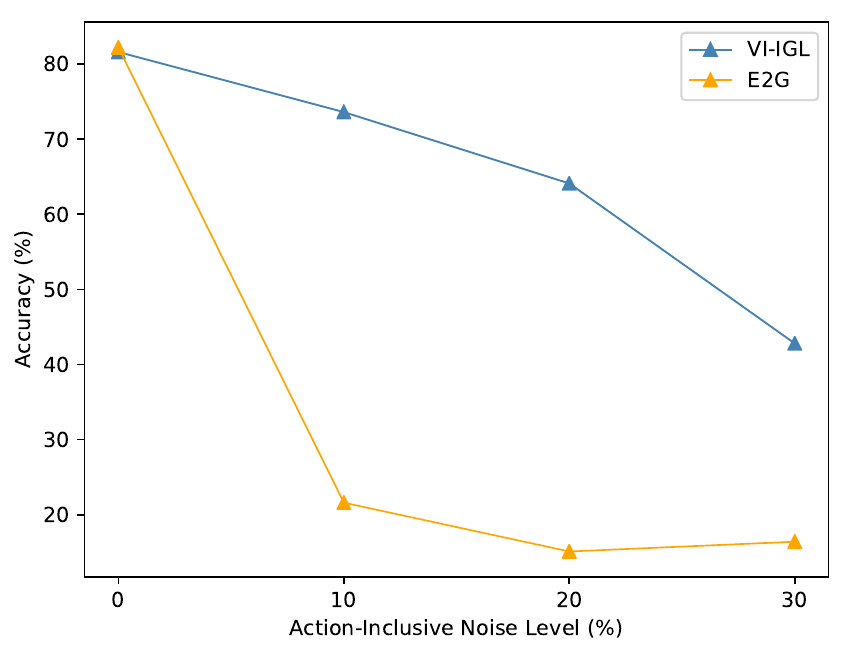}} 
    \subfigure[Context-inclusive]{\includegraphics[width=0.35\textwidth]{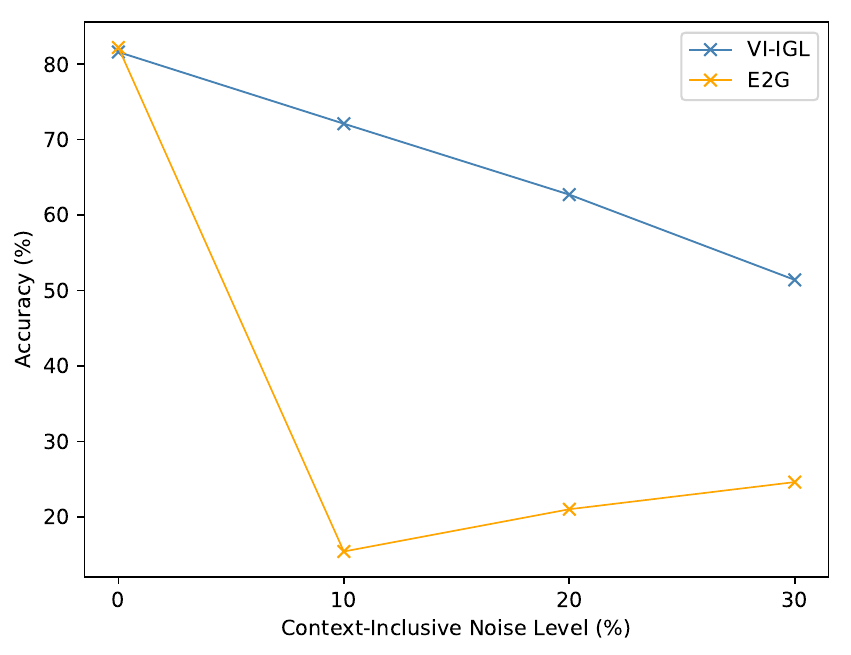}}
    \subfigure[Context-action-inclusive]{\includegraphics[width=0.35\textwidth]{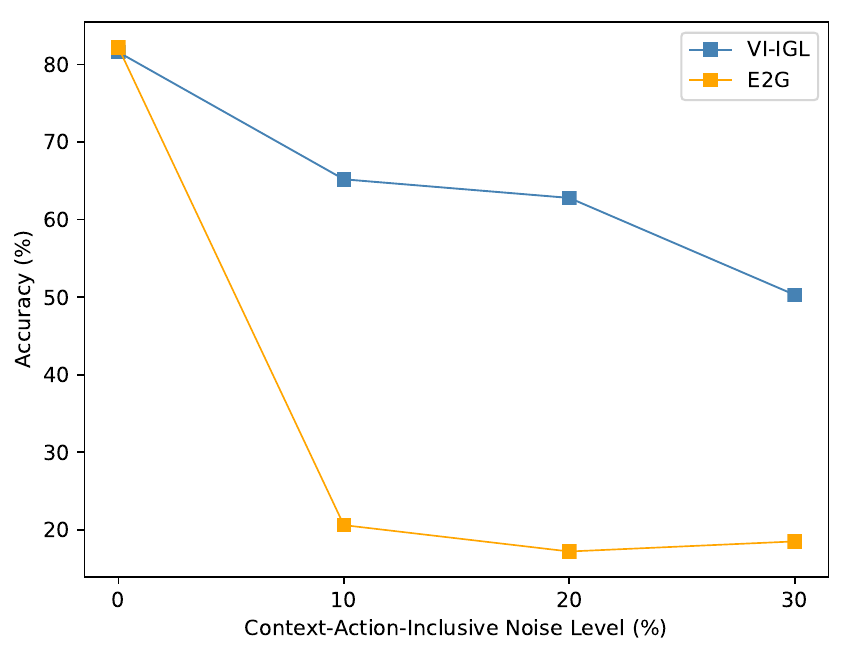}}
\caption{Policy accuracy under different noise level: Our VI-IGL algorithm outperforms batch \textsf{E2G}~\citep{pmlr-v139-xie21e} in all noisy environments and across all noise levels. The results are averaged over 16 trials.}
\label{fig:1}
\end{figure}

\textbf{Why previous IGL method fails.} Recall that solving an appropriate reward decoder in the previous IGL method is given by~\citep[Assumption 2]{pmlr-v139-xie21e}, which states that there exists a reward decoder that well distinguishes between the feedback (distribution) generated from a latent reward of 0 and the one generated from a latent reward of 1. When additional noises present in the feedback, these two distributions can be quite similar. For example, for context-inclusive noises, a latent reward of 0 can also generate an image of digit ``1''~($l_{x_t}=4$ and $r_t=0$). Hence, the condition easily fails and the performance degrades.

\subsection{Necessity of Regularization}
\label{sec: 6.2}

In this section, we show that including the regularization term $I(X,A;R_\psi)$ in Objective~(\ref{obj}) helps achieve a more consistent algorithm performance.  We compare the algorithm performance when optimizing the unregularized objective~($\beta=0$) and the regularized objective~($\beta=10$). Note that the case of $\beta=0$ corresponds to minimizing only $I(Y;X,A|R_\psi)$. The results are summarized in Table~\ref{tab_112}.~(Results for other selections of $\beta$ can be found in Appendix~\ref{sec:F.2}.) The results show that regularization significantly improves the performance. 

\begin{table}[h]
\begin{center}
\begin{tabular}{ |l||c|c|c| }
\hline
\text{Methods}&
VI-IGL ($\beta=0$) & VI-IGL ($\beta=10$)\\
\hline
\hline
\texttt{I} & $60.6\pm15.5$ &  $\bm{71.0\pm16.2}$  \\
\hline
\texttt{A} & $64.8\pm17.9$ &  $\bm{73.6\pm16.0}$   \\
\hline
\texttt{C} & $52.4\pm25.3$ & $\bm{72.1\pm10.2}$  \\
\hline
\texttt{C-A} & $63.4\pm16.4$ & $\bm{65.2\pm16.3}$  \\
\hline
\hline
\texttt{N} & $55.7\pm27.3$ & $\bm{81.6\pm7.9}$  \\
\hline
\end{tabular}
\caption{Noise level=0.1. The results are averaged over 16 trials~(\texttt{I}: independent, \texttt{A}: action-inclusive, \texttt{C}: context-inclusive, \texttt{C-A}: context-action-inclusive, \texttt{N}: noiseless setting).}
\label{tab_112}
\end{center}
\end{table}

\subsection{Ablation Experiments}
\label{sec: 6.3}

\paragraph{6.3.1. Selection of $f$-divergences.} Recall that in Objective~(\ref{obj-f}), we use $f_1$ and $f_2$ as general measures of $I(Y;X,A|R_\psi)$ and $I(X,A;R)$, respectively. We analyze how the selection of $f$-divergences affects the performance. We test three pairs of $f_1$-$f_2$: (i) KL-KL: both $f_1$ and $f_2$ are KL divergence, i.e., $f_1(x)=f_2(x)=x\log x$~(this case corresponds to Objective~(\ref{obj})), (ii) $\chi^2$-$\chi^2$: both $f_1$ and $f_2$ are Pearson-$\chi^2$ divergence, i.e., $f_1(x)=f_2(x)=(x-1)^2$, and (iii) $\chi^2$-KL: $f_1(x)=x\log x$ is KL divergence and $f_2(x)=(x-1)^2$ is Pearson-$\chi^2$ divergence. Note that in the last case, the objective value, i.e., $I_{\chi^2}(Y;X,A|R_\psi)-\beta\cdot I(X,A;R_\psi)$, upper bounds the value of Objective~(\ref{obj}).\footnote{By the inequality $\log\le x-1$, we have that $D_\textup{KL}(\sP\|\sQ)=\E_\sP[\log(\frac{\textup{d}\sP}{\textup{d}\sQ})]\le\E_\sP[(\frac{\textup{d}\sP}{\textup{d}\sQ}-1)]=\E_{\sQ}[(\frac{\textup{d}\sP}{\textup{d}\sQ})^2]-1=D_\chi^2(\sP\|\sQ)$.} We summarize the results in Table~\ref{tab_4} for a feedback-dependent reward decoder and $\beta=10$. The results show that different $f$-divergences benefit from different types of noises. 

\begin{table}[!h]
\begin{center}
\begin{tabular}{ |l||c|c|c| }
\hline
$f_1$-$f_2$ & KL-KL & $\chi^2$-$\chi^2$ & $\chi^2$-KL \\
\hline
\texttt{I}    & $71.0\pm16.2$ & $72.7\pm17.4$  & $\bm{74.1\pm12.7}$  \\
\hline
\texttt{A}    & $\bm{73.6\pm16.0}$ & $65.3\pm11.1$ & $71.5\pm16.7$  \\
\hline
\texttt{C}    &  $72.1\pm10.2$ & $\bm{76.2\pm11.5}$ & $69.4\pm11.5$  \\
\hline
\texttt{C-A}    &  $65.2\pm16.3$  & $\bm{70.4\pm15.5}$ & $64.6\pm14.7$ \\
\hline
\hline
\texttt{N}  & $\bm{81.6\pm7.9}$  & $74.8\pm13.3$ & $77.3\pm11.1$ \\
\hline
\end{tabular}
\caption{Selection of $f$-divergences: The results are averaged over 16 trials~(\texttt{I}: independent, \texttt{A}: action-inclusive, \texttt{C}: context-inclusive, \texttt{C-A}: context-action-inclusive, \texttt{N}: noiseless setting).}
\label{tab_4}
\end{center}
\end{table}

\paragraph{6.3.2. Input of reward decoder.} We empirically analyze how the input of the reward decoder affects the actual performance. Particularly, we consider two types of input: (i) feedback $Y$ and (ii) context-action-feedback $(X,A,Y)$. We present the results in Table~\ref{tab_2} for $\beta=10$ and KL-KL divergence measure. The results show that in all cases, using a feedback-dependent reward decoder class leads to better performance than a context-action-feedback-dependent reward decoder class. 

\begin{table}[!h]
\begin{center}
\begin{tabular}{ |l||c|c| }
\hline
\text{Input} & $Y$ & $(X,A,Y)$ \\
\hline
\hline
\texttt{I}    & $\bm{71.0\pm16.2}$ & $42.0\pm24.1$  \\
\hline
\texttt{A}    & $\bm{73.6\pm16.0}$ & $49.6\pm24.4$  \\
\hline
\texttt{C}    &  $\bm{69.3\pm10.9}$ &  $46.2\pm16.9$ \\
\hline
\texttt{C-A}    & $\bm{72.1\pm10.2}$  & $59.4\pm17.4$  \\
\hline
\hline
\texttt{N} & $\bm{81.6\pm7.9}$ & $62.5\pm19.1$ \\
\hline
\end{tabular}
\caption{Input of Reward Decoder: The results are averaged over 16 trials~(\texttt{I}: independent, \texttt{A}: action-inclusive, \texttt{C}: context-inclusive, \texttt{C-A}: context-action-inclusive, \texttt{N}: noiseless setting).}
\label{tab_2}
\end{center}
\end{table}

%% file: conclusion.tex

Regarding the limitations of our methodology and analysis, 
we observed that the variance of the numerical performance could be considerably large in some experiments. We hypothesize that this issue could be related to jointly training multiple networks and model initialization, also reported by Xie~et~al.~\citeyearpar{pmlr-v139-xie21e}, which could be an interesting topic for future studies.
Also, an extension of our analysis is to relax the full conditional independence assumption~(Assumption~\ref{as_fci}). Such an extension could follow \cite{xie2022interactiongrounded}'s idea on the Action-Inclusive IGL~(AI-IGL), where the feedback may also be affected by the action.

%% file: Appendix.tex
\allowdisplaybreaks

\section{Proof of Theorem~\ref{thm_vi-igl}: The VI-IGL Optimization Problem}
\label{p_thm_vi-igl}

\begin{proof}
    The theorem is a direct application of Proposition~\ref{prop_vrkl}. The optimization problem possesses three levels: (i) the inner level minimizes over function class $T\in\gT$ to estimate $I(X,A;R_\psi)=D_\textup{KL}(\mathbb{P}_{XAR_\psi}||\mathbb{P}_{XA}\otimes\mathbb{P}_{R_\psi})=\sup_{T\in\mathcal{T}}\{\mathbb{E}_{\mathbb{P}_{XAR_\psi}}[T]-\mathbb{E}_{\mathbb{P}_{XA}\otimes\mathbb{P}_{R_\psi}}[e^T]\}$, (ii) the medium level maximizes over function class $G\in\gG$ to estimate $I(Y;X,A|R_\psi)=D_\textup{KL}(\mathbb{P}_{XAYR_\psi}||\mathbb{P}_{Y|R_\psi}\otimes\mathbb{P}_{XAR_\psi})=\sup_{G\in\mathcal{G}}\{\mathbb{E}_{\mathbb{P}_{XAYR_\psi}}[G]-\mathbb{E}_{\mathbb{P}_{Y|R_\psi}\otimes\mathbb{P}_{XAR_\psi}}[e^G]\}$, and (iii) the outer level finds the desired reward decoder. 
\end{proof}

\section{Proof of Theorem~\ref{thm_sc}: Sample Complexity of VI-IGL Optimization Problem}
\label{p_thm_sc}

Recall that the optimization problem~(\ref{opt_KL}) is minimizing
\begin{equation}
\label{eq_20}
\begin{aligned}
    \gL(\psi)=\max_{G\in\gG}\min_{T\in\gT}\Big\{&\E_{\sP_{XAYR_\psi}}[G]-\E_{\sP_{Y|R_\psi}\otimes\sP_{XA|R_\psi}}\left[e^{G}\right]- \beta\cdot\left(\E_{\sP_{XAR_\psi}}[T]-\E_{\sP_{XA}\otimes\sP_{R_\psi}}\left[e^{T}\right]\right)\Big\}
\end{aligned}
\end{equation}
over the reward decoder $\psi\in\Psi$. In the offline setting, the learner has access to a $K$-size dataset $\gD=\{(x_k,a_k,y_k)\}_{k=1}^K$ collected by the behavior policy $\pi_\textup{b}:\gX\to\Delta_\gA$. Particularly, at round $k$, a context $x_k\sim d_0$ is drawn from the context distribution. The behavior policy returns $a_k\sim\pi_\textup{b}(\cdot|x_k)$ and receives feedback $y_k$ from the environment. 

The algorithm constructs the empirical objective $\widehat{\gL}(\psi)$ from the dataset for any reward decoder $\psi\in\Psi$ and outputs the minimizer $\widehat{\psi}=\arg\min_{\psi\in\Psi}\widehat{\gL}(\psi)$. To show Theorem~\ref{thm_sc}, it suffices to show that
\begin{equation*}
\begin{aligned}
    \left|\gL(\psi)-\widehat{\gL}(\psi)\right|\le \epsilon+ \max\{1,\beta\}\cdot O\left(\frac{(1-c)^2}{c^2}\cdot \sum_{y\in\gY^{\epsilon}_{\gF}:\sigma_{\pi_\textup{b}}^\epsilon(y)>0}\sqrt{\frac{e^{2B}+B^2}{K\cdot \sigma_{\pi_\textup{b}}^\epsilon(y)}\log\left(\frac{|\gY^{\epsilon}_{\gF}|d_{\Psi,\gT,\gG}}{\delta}\right)}\right)
\end{aligned}
\end{equation*}
for any reward decoder $\psi\in\Psi$ with high probability, where the parameters $\gY^{\epsilon}_{\gF},\sigma_{\pi_\textup{b}}^\epsilon$, and $d_{\Psi,\gT,\gG}$ are specified in the proof. Once obtained, we set the parameter $\epsilon=K^{-1/2}$ and invoke the following inequality
\begin{equation}
    \gL(\widehat{\psi})-\gL(\psi^*)\le \left|\gL(\widehat{\psi})-\widehat{\gL}(\widehat{\psi})\right|+\underbrace{\widehat{\gL}(\widehat{\psi})-\widehat{\gL}(\psi^*)}_\text{$\le 0$}+\left|\widehat{\gL}(\psi^*)-\gL(\psi^*)\right|
\end{equation}
to conclude the proof. Since the optimization problem over function classes $\gG$ and $\gT$ are decoupled, we define
\begin{gather}
    \gL_1(\psi)=\max_{G\in\gG}\left\{\E_{\sP_{XAYR_\psi}}[G]-\E_{\sP_{Y|R_\psi}\otimes\sP_{XA|R_\psi}}\left[e^G\right]\right\}\\
    \gL_2(\psi)=\max_{T\in\gT}\left\{\E_{\sP_{XAR_\psi}}[T]-\E_{\sP_{XA}\otimes\sP_{R_\psi}}\left[e^T\right]\right\}
\end{gather}
Hence, we have that $\gL(\psi)=\gL_1(\psi)-\beta\cdot \gL_2(\psi)$. 

The details of the algorithm is given as follows. We consider a feedback-dependent reward decoder class, where $\psi(y)\in[c,1-c]$ is the decoded probability given by $\psi\in\Psi$ that the feedback $y\in\gY$ is associated with a latent reward of 1. For convenience, we define $\psi_1(y):=\psi(y)$ and $\psi_0(y):=1-\psi(y)$, where the subscript is decoded binary reward. The algorithm computes the empirical counterpart of $\gL_1(\psi)$ and $\gL_2(\psi)$ as follows.
\begin{gather}
    \widehat{\gL}_1(\psi)=\max_{G\in\gG}\Bigg\{\frac{1}{K}\sum_{k=1}^K\sum_{r=0,1}\psi_r(y_k)\cdot \Big(G(x_k,a_k,y_k,r)-\E_{\widehat{\sP}_{Y^\epsilon|R_\psi=r}}\left[e^{G(x_k,a_k,\tilde{y},r)}\right]\Big)\Bigg\}\\
    \widehat{\gL}_2(\psi)=\max_{T\in\gT}\Bigg\{\frac{1}{K}\sum_{k=1}^K \sum_{r=0,1}\left(\psi_r(y_k)\cdot T(x_k,a_k,r)-\widehat{p}_\psi(r)\cdot e^{T(x_k,a_k,r)}\right)\Bigg\}
\end{gather}
where $\tilde{y}\sim\widehat{\sP}_{Y^\epsilon|R_\psi=r}$ in $\widehat{\gL}_1(\psi)$ is the empirical estimation of $\sP_{Y|R_\psi=r}$ constructed from the dataset (see details in the proof) and $\widehat{p}_\psi(r):=\frac{1}{K}\sum_{k=1}^K\psi_1(y_k)$. In the proof, we aim to bound the estimation errors $|\gL_1(\psi)-\widehat{\gL}_1(\psi)|$ and $|\gL_2(\psi)-\widehat{\gL}_2(\psi)|$.

\begin{proof}
Fix a reward decoder $\psi\in\Psi$.

\paragraph{Step 1. Bounding $|\gL_2(\psi)-\widehat{\gL}_2(\psi)|$.}

Recall that $\gT$ is bounded by $B$. Hence, with probability at least $1-\delta$ and applying a union bound over the function classes $\gT$, the estimation errors are bounded by
\begin{gather}
    \left|\E_{\sP_{XAR_\psi}}[T]-\frac{1}{K}\sum_{k=1}^K\sum_{r=0,1}\psi_r(y_k)\cdot T(x_k,a_k,r)\right|\le O\left(\sqrt{\frac{B^2}{K}\log\left(\frac{d_{\gT}}{\delta}\right)}\right)\label{ieq_35}\\
    \left|\E_{\sP_{XA}\otimes\sP_{R_\psi}}\left[e^T\right]-\frac{1}{K}\sum_{k=1}^K\sum_{r=0,1}\widehat{p}_\psi(r)\cdot e^{T(x_k,a_k,r)}\right|\le O\left(\sqrt{\frac{e^{2B}}{K}\log\left(\frac{d_{\gT}}{\delta}\right)}\right)
\end{gather}
for any function $T\in\gT$, where $d_{\gT}$ is the statistical complexity of the function class $\gT$. Particularly, if $\gT$ is finite, we have that $d_{\gT}=|\gT|$. 

\paragraph{Step 2. Bounding $|\gL_1(\psi)-\widehat{\gL}_1(\psi)|$.} Fix a function $G\in\gG$. Recall that $G$ is bounded by $B$. Hence, with probability at least $1-\delta$, we have that
\begin{equation}
\label{ieq_33}
    \left|\E_{\sP_{XAYR_\psi}}[G]-\frac{1}{K}\sum_{k=1}^K\sum_{r=0,1}\psi_r(y_k)\cdot G(x_k,a_k,y_k,r)\right|\le O\left(\sqrt{\frac{B^2}{K}\log\left(\frac{1}{\delta}\right)}\right)
\end{equation}
The challenge is to analyze the estimation error 
\begin{equation}
\label{eq_71}
    \left|\E_{\sP_{Y|R_\psi}\otimes\sP_{XAR_\psi}}\left[e^G\right]-\frac{1}{K}\sum_{k=1}^K\sum_{r=0,1}\psi_r(y_k)\cdot\E_{\widehat{\sP}_{Y^\epsilon|r}}\left[e^{G(x_k,a_k,\tilde{y},r)}\right]\right|
\end{equation}
To handle the continuous feedback space, we first introduce the notion of $\epsilon$-covering, which results in a finite ``clusterings'' of the feedback and yields nice statistical properties.

\begin{definition}[$\epsilon$-covering]
\label{def_cov}
Let $\gG\subset\{\gY\to\sR\}$ denote a function class. A (finite) set $\gY^\epsilon_{\gG}\subset\gY$ is said to be $\epsilon$-covering the space $\gY$ with respect to function class $\gG$ if for any $y\in\gY$, there exists $y^\epsilon\in\gY_\gG^\epsilon$ such that $\max_{G\in\gG}|G(y)-G(y^\epsilon)|\le\epsilon$. Further, we denote by $|\gY^\epsilon_{\gG}|$ the $\epsilon$-covering number.
\end{definition}

\begin{remark}
Definition~\ref{def_cov} is a classic $\epsilon$-covering of space $\gY$ equipped with (pseudo-)metric $\rho(y,y')=\max_{G\in\gG}|G(y)-G(y')|$.\footnote{To show $\rho$ is indeed a metric, note that 1) $\rho(y,y')=\rho(y',y)\ge 0$ $\rho(y,y)=0$ for any $y,y'\in\gY$ and 2) $\rho(y,y')\le\rho(y,y'')+\rho(y'',y')$ for any $y,y',y''\in\gY$.} For example, if the class $\gG$ includes $\alpha$-Lipschitz functions, i.e., $|G(y)-G(y')|\le\alpha\cdot\|y-y'\|_2$, then $\gY_\gG^\epsilon$ is an $(\frac{\epsilon}{\alpha})$-covering of $\gY$ equipped with metric $\rho(y,y')=\|y-y'\|_2$.
\end{remark}

In the following, we denote by $\gY^{\epsilon}_{\gF}$ an $\epsilon$-covering with respect to the joint function class $\gF:=\Psi\cup\{e^{G(x,a,\cdot,r)}:(x,a,r)\in\gX\times\gA\times\{0,1\}\}_{G\in\gG}$ and let $s:\gY\mapsto\gY_{\gF}^\epsilon$ be a mapping from any $y\in\gY$ to $\gY_{\gF}^\epsilon$ such that $\max_{F\in\gF}|F(y)-F(y^\epsilon)|\le\epsilon$. Let $\sigma_{\pi_\textup{b}}\in\Delta_\gY$ be the feedback distribution induced by the behavior policy $\pi_\textup{b}$. We denote by $\sigma_{\pi_\textup{b}}^\epsilon$ the corresponding distribution on the $\epsilon$-covering $\gY^{\epsilon}_{\gF}$. Specifically, the mass at any $y^\epsilon\in\gY^{\epsilon}_{\gF}$ is given by $\sigma_{\pi_\textup{b}}^\epsilon(y^\epsilon):=\int_{y:s(y)=y^\epsilon}\textup{d}\sigma_{\pi_\textup{b}}(y)$. Further, the reward decoder $\psi\in\Psi$ induces posterior distributions $\sP_{Y|R_\psi}$ and $\sP_{Y^\epsilon|R_\psi}$ conditioned to the decoded reward on $\gY$ and $\gY^{\epsilon}_{\gF}$, respective. By Definition~\ref{def_cov}, the expectation with respect to the distribution $\sP_{Y|R_\psi}$ can be well estimated by the expectation computed from $\sP_{Y^\epsilon|R_\psi}$.

\textbf{Sub-Step 2.1. Construction of $\widehat{\sP}_{Y^\epsilon|r}$.}\qquad The construction of $\widehat{\sP}_{Y^\epsilon|r}$, which involves: 1) computing the empirical feedback distribution $\sigma_{\pi_\textup{b}}^\epsilon:=\frac{1}{K}\sum_{k=1}^K\mathbbm{1}[s(y)=y^\epsilon]$ for any $y^\epsilon\in\gY^{\epsilon}_{\gF}$, and 2) utilizing Bayes rules to estimate the posterior distribution by
\begin{equation}
    \widehat{\sP}_{Y^\epsilon|R_\psi=r}(y^\epsilon):=\frac{\widehat{\sigma}_{\pi_\textup{b}}^\epsilon(y^\epsilon)\cdot\psi_r(y^\epsilon)}{\sum_{y\in\gY^{\epsilon}_{\gF}}\widehat{\sigma}_{\pi_\textup{b}}^\epsilon(y)\cdot\psi_r(y)}
\end{equation}
for any $y^\epsilon\in\gY^{\epsilon}_{\gF}$. Hence, the error~(\ref{eq_71}) can be further written as
\begin{equation}
\label{ieq_71}
\begin{aligned}
    &\left|\E_{\sP_{Y|R_\psi}\otimes\sP_{XAR_\psi}}\left[e^G\right]-\frac{1}{K}\sum_{k=1}^K\sum_{r=0,1}\psi_r(y_k)\cdot\E_{\widehat{\sP}_{Y^\epsilon|R_\psi}}\left[e^{G(x_k,a_k,\tilde{y}_r,r)-1}\right]\right|\\
    \le & \underbrace{\left|\E_{\sP_{Y|R_\psi}\otimes\sP_{XAR_\psi}}\left[e^G\right]-\E_{\sP_{Y^\epsilon|R_\psi}\otimes\sP_{XAR_\psi}}\left[e^G\right]\right|}_\text{$\le\epsilon$} \\
    &+\underbrace{\left|\E_{\sP_{Y^\epsilon|R_\psi}\otimes\sP_{XAR_\psi}}\left[e^G\right]-\frac{1}{K}\sum_{k=1}^K\sum_{r=0,1}\psi_r(y_k)\cdot\E_{\sP_{Y^\epsilon|R_\psi=r}}\left[e^G\right]\right|}_\text{concentration error}\\
    &+ \left|\frac{1}{K}\sum_{k=1}^K\sum_{r=0,1}\psi_r(y_k)\cdot\left(\E_{\sP_{Y^\epsilon|R_\psi=r}}\left[e^{G(x_k,a_k,\tilde{y},r)}\right]-\E_{\widehat{\sP}_{Y^\epsilon|R_\psi=r}}\left[e^{G(x_k,a_k,\tilde{y},r)}\right]\right)\right|
\end{aligned}
\end{equation}
Observe that the last term is bounded by
\begin{equation*}
\label{ieq_55}
\begin{aligned}
    \max_{r}\left\|\sP_{Y^\epsilon|R_\psi=r}-\widehat{\sP}_{Y^\epsilon|R_\psi=r}\right\|_1 \cdot e^{B-1}
\end{aligned}
\end{equation*}
for any $(\psi,G)\in\Psi\times\gG$ and $(x,a)\in\gX\times\gA$. It remains to bound the error $\|\sP_{Y^\epsilon|R_\psi=r}-\widehat{\sP}_{Y^\epsilon|R_\psi=r}\|_1$. 

\textbf{Sub-Step 2.2. Bounding $\|\sP_{Y^\epsilon|R_\psi=r}-\widehat{\sP}_{Y^\epsilon|R_\psi=r}\|_1$.}\qquad To start with, by~\citep[Lemma A.1]{xie2021policy}, with probability at least $1-\delta$ and applying union bound over $\gY^{\epsilon}_{\gF}$, it holds that
\begin{equation*}
    \left|\widehat{\sigma}_{\pi_\textup{b}}^\epsilon(y^\epsilon)-\sigma_{\pi_\textup{b}}^\epsilon(y^\epsilon)\right|\le O\left(\sqrt{\frac{1}{K\cdot \sigma_{\pi_\textup{b}}^\epsilon(y^\epsilon)}\log\left(\frac{|\gY^{\epsilon}_{\gF}|}{\delta}\right)}\right)
\end{equation*}
for any $y^\epsilon\in\gY^{\epsilon}_{\gF}$. Recall that $\psi_r$ is bounded between $[c,1-c]$ where $0<c<\frac{1}{2}$. We have that
\begin{align*}
    &\left|\sP_{Y^\epsilon|r}(y^\epsilon)-\widehat{\sP}_{Y^\epsilon|r}(y^\epsilon)\right|\\
    =&\left|\frac{\widehat{\sigma}_{\pi_\textup{b}}^\epsilon(y^\epsilon)\cdot\psi_r(y^\epsilon)}{\sum_{y\in\gY^{\epsilon}_{\gF}}\widehat{\sigma}_{\pi_\textup{b}}^\epsilon(y)\cdot\psi_r(y)}-\frac{\sigma_{\pi_\textup{b}}^\epsilon(y^\epsilon)\cdot\psi_r(y^\epsilon)}{\sum_{y\in\gY^{\epsilon}_{\gF}}\sigma_{\pi_\textup{b}}^\epsilon(y)\cdot\psi_r(y)}\right|\\
    = & \left|\frac{\left(\widehat{\sigma}_{\pi_\textup{b}}^\epsilon(y^\epsilon)-\sigma_{\pi_\textup{b}}^\epsilon(y^\epsilon)\right)\cdot\psi_r(y^\epsilon)}{\sum_{y\in\gY^{\epsilon}_{\gF}}\widehat{\sigma}_{\pi_\textup{b}}^\epsilon(y)\cdot\psi_r(y)}+\frac{\sigma_{\pi_\textup{b}}^\epsilon(y^\epsilon)\cdot\psi_r(y^\epsilon)}{\sum_{y\in\gY^{\epsilon}_{\gF}}\widehat{\sigma}_{\pi_\textup{b}}^\epsilon(y)\cdot\psi_r(y)}-\frac{\sigma_{\pi_\textup{b}}^\epsilon(y^\epsilon)\cdot\psi_r(y^\epsilon)}{\sum_{y\in\gY^{\epsilon}_{\gF}}\sigma_{\pi_\textup{b}}^\epsilon(y)\cdot\psi_r(y)}\right|\\
    \le & O\left(\frac{1-c}{c}\sqrt{\frac{1}{K\cdot \sigma_{\pi_\textup{b}}^\epsilon(y^\epsilon)}\log\left(\frac{|\gY^{\epsilon}_{\gF}|}{\delta}\right)}+\frac{(1-c)^2}{c^2}\cdot \sum_{y\in\gY^{\epsilon}_{\gF}}\sqrt{\frac{1}{K\cdot \sigma_{\pi_\textup{b}}^\epsilon(y)}\log\left(\frac{|\gY^{\epsilon}_{\gF}|}{\delta}\right)}\right)\\
    \le & O\left(\frac{(1-c)^2}{c^2}\cdot \sum_{y\in\gY^{\epsilon}_{\gF}:\sigma_{\pi_\textup{b}}^\epsilon(y)>0}\sqrt{\frac{1}{K\cdot \sigma_{\pi_\textup{b}}^\epsilon(y)}\log\left(\frac{|\gY^{\epsilon}_{\gF}|}{\delta}\right)}\right)
\end{align*}
where the second inequality holds by the fact that $\sum_{y\in\gY^{\epsilon}_{\gF}}\widehat{\sigma}_{\pi_\textup{b}}^\epsilon(y)\cdot\psi_r(y)$ and $\sum_{y\in\gY^{\epsilon}_{\gF}}\sigma_{\pi_\textup{b}}^\epsilon(y)\cdot\psi_r(y)$ are bounded between $[c,1-c]$. Hence, for any $r\in\{0,1\}$, it holds that
\begin{equation}
\label{ieq_119}
    \left\|\sP_{Y^\epsilon|R_\psi=r}-\widehat{\sP}_{Y^\epsilon|R_\psi=r}\right\|_1\le O\left(\frac{(1-c)^2}{c^2}\cdot \sum_{y\in\gY^{\epsilon}_{\gF}:\sigma_{\pi_\textup{b}}^\epsilon(y)>0}\sqrt{\frac{|\gY^{\epsilon}_{\gF}|^2}{K\cdot \sigma_{\pi_\textup{b}}^\epsilon(y)}\log\left(\frac{|\gY^{\epsilon}_{\gF}|}{\delta}\right)}\right)
\end{equation}
Therefore, combining Inequalities (\ref{ieq_71})$\sim$(\ref{ieq_119}) and applying a union bound over $G\in\gG$ yields,
\begin{equation}
\label{ieq_129}
\begin{aligned}
    &\left|\E_{\sP_{Y|R_\psi}\otimes\sP_{XAR_\psi}}\left[e^G\right]-\frac{1}{K}\sum_{k=1}^K\sum_{r=0,1}\psi_r(y_k)\cdot\E_{\widehat{\sP}_{Y^\epsilon|r}}\left[e^{G(x_k,a_k,\tilde{y},r)}\right]\right|\\
    \le & O\left(\frac{(1-c)^2}{c^2}\cdot \sum_{y\in\gY^{\epsilon}_{\gF}:\sigma_{\pi_\textup{b}}^\epsilon(y)>0}\sqrt{\frac{e^{2B}|\gY^{\epsilon}_{\gF}|^2}{K\cdot \sigma_{\pi_\textup{b}}^\epsilon(y)}\log\left(\frac{|\gY^{\epsilon}_{\gF}|d_\gG}{\delta}\right)}\right)
\end{aligned}
\end{equation}
where $d_{\gG}$ is the statistical complexity of the function class $\gG$, with $d_{\gG}=|\gG|$ for finite class $\gG$.

\paragraph{Step 3. Putting everything together.} 

Combining Inequalities (\ref{ieq_35})$\sim$(\ref{ieq_33}) and (\ref{ieq_129}) and applying a union bound over the reward decoder class $\Psi$, we have that
\begin{equation*}
\begin{aligned}
    \left|\gL(\psi)-\widehat{\gL}(\psi)\right|\le \epsilon+ \max\{1,\beta\}\cdot O\left(\frac{(1-c)^2}{c^2}\cdot \sum_{y\in\gY^{\epsilon}_{\gF}:\sigma_{\pi_\textup{b}}^\epsilon(y)>0}\sqrt{\frac{(e^{2B}+B^2)|\gY^{\epsilon}_{\gF}|^2}{K\cdot \sigma_{\pi_\textup{b}}^\epsilon(y)}\log\left(\frac{|\gY^{\epsilon}_{\gF}|d_{\Psi,\gT,\gG}}{\delta}\right)}\right)
\end{aligned}
\end{equation*}
for any $\psi\in\Psi$, where we denote by $d_{\Psi,\gT,\gG}$ the statistical complexity of the joint function classes $\Psi$, $\gT$, and $\gG$, with $d_{\Psi,\gT,\gG}=|\Psi||\gT||\gG|$ for finite classes. Define the capacity number
\begin{equation}
\label{eq_capa}
    \gC(\gY_\gP^\epsilon,B):=\sum_{y\in\gY^{\epsilon}_{\gF}:\sigma_{\pi_\textup{b}}^\epsilon(y)>0}\frac{(e^{2B}+B^2)|\gY^{\epsilon}_{\gF}|^3}{\sigma_{\pi_\textup{b}}^\epsilon(y)}
\end{equation}
By Cauchy-Schwartz inequality, we further have
\begin{equation*}
    \left|\gL(\psi)-\widehat{\gL}(\psi)\right|\le \epsilon+ \max\{1,\beta\}\cdot O\left(\frac{(1-c)^2}{c^2}\cdot\sqrt{\frac{\gC(\gY_\gP^\epsilon,B)}{K}\log\left(\frac{|\gY^{\epsilon}_{\gF}|d_{\Psi,\gT,\gG}}{\delta}\right)} \right)
\end{equation*}

Set $\epsilon=K^{-1/2}$ and we conclude the proof.
\end{proof}

\section{Proof of Theorem~\ref{thm_4}: Regularization (Almost) Ensures Conditional Independence}
\label{p_thm_4}

\begin{proof}
Under the realizability assumption, there exists either (i) a context-action-dependent reward decoder $\tilde{\psi}:\gX\times\gA\to[0,1]$ or (ii) a feedback-dependent reward decoder $\Tilde{\psi}:\gY\to[0,1]$ such that $I(Y;X,A|R_{\tilde{\psi}})=0$.\footnote{Note that if a reward decoder depends on the $(X,A,Y)$ tuple, it can be regarded as case (i).} We first show that 
\begin{equation}
\label{eq_141}
    I(X,A;R_{\tilde{\psi}})=I(Y;X,A)=I(Y;R)
\end{equation}
holds for both cases, where $R$ is the true latent reward. 

\textbf{Case (i). } Note that by the chain rules of CMI, we derive
\begin{equation*}
    0=I(Y;X,A|R_{\Tilde{\psi}})=I(Y;X,A)+\underbrace{I(Y;R_{\Tilde{\psi}}|X,A)}_\text{$=0$}-I(Y;R_{\Tilde{\psi}})
\end{equation*}
where the second term $I(Y;R_{\Tilde{\psi}}|X,A)$ on the RHS is zero as $R_\psi$ is context-action-dependent. Hence, we have that $I(Y;R_{\tilde{\psi}})=I(Y;X,A)$. Further, note that the following Markov chain holds:
\begin{equation*}
    (X,A)\rightarrow R_{\Tilde{\psi}}\rightarrow Y
\end{equation*}
By the data processing inequality, we derive that $I(X,A;R_{\Tilde{\psi}})\le I(Y;X,A)$ and the equality holds \emph{if and only if} $I(Y;R_{\tilde{\psi}})=I(Y;X,A)$. Therefore, we have that $I(X,A;R_{\Tilde{\psi}})=I(Y;X,A)$. Since the true latent reward $R$ is context-action-dependent, following the same analysis, we have
\begin{equation}
\label{eq_156}
    I(Y;X,A)=I(Y;R)=I(X,A;R)   
\end{equation}
Combining the analysis above, Equation~(\ref{eq_141}) is proved.

\textbf{Case (ii). } Note that the following Markov chain holds for any feedback-dependent reward decoder:
\begin{equation*}
    (X,A)\rightarrow Y\rightarrow R_{\Tilde{\psi}}
\end{equation*}
Then, the data processing inequality implies that $I(X,A;R_{\Tilde{\psi}})\le I(Y;X,A)$ and the equality holds \emph{if and only if} $I(Y;X,A|R_{\Tilde{\psi}})=0$. Therefore, we derive $I(X,A;R_{\Tilde{\psi}})=I(Y;X,A)=I(Y;R)$, where the last equality holds by Equation~(\ref{eq_156}). This also implies that for feedback-dependent reward decoder class, it holds that
\begin{equation*}
    \min_{\psi\in\Psi}\{I(Y;X,A|R_\psi)-\beta\cdot I(X,A;R_\psi)\}\ge -\beta\cdot I(Y;R)
\end{equation*}
Therefore, when $\Psi$ is feedback-dependent, a reward decoder $\psi:\gY\to[0,1]$ attains the minimum if and only if $I(Y;X,A|R_\psi)=0$.

Once Equation~(\ref{eq_141}) is obtained, we have that
\begin{equation*}
    \min_{\psi\in\Psi}\{I(Y;X,A|R_\psi)-\beta\cdot I(X,A;R_\psi)\}\le I(Y;X,A|R_{\tilde{\psi}})-\beta\cdot I(X,A;R_{\tilde{\psi}})=-\beta\cdot I(Y;R)
\end{equation*}
Let $\psi^*$ denote any reward decoder optimizing Objective~(\ref{obj}). Rearranging the terms proves
\begin{equation*}
    I(Y;X,A|R_{\psi^*})\le\beta\cdot(I(X,A;R_{\psi^*})-I(Y;R))\le \beta\cdot(\log2-I(Y;R))
\end{equation*}
where the second inequality holds by the fact that $R_\psi^*$ is a 0-1 random variable. Therefore, we conclude the proof.
\end{proof}

\section{The Variational Representation of $f$-divergences}
\label{sec_var}

\begin{proposition}[Variational representation of $f$-divergences~\citep{10.1109/TIT.2010.2068870}]
\label{prop_varrep}
Let $f:\sR_+\mapsto\sR$ be a convex, lower-semicontinuous function satisfying $f(1)=0$. Consider $\sP,\sQ\in\Delta_\gS$ as two probability distributions on space $\gS$. Then,
\begin{equation*}
\begin{aligned}
    D_f(\sP\|\sQ)=&\E_{\sQ}\left[f\left(\frac{d\sP}{d\sQ}\right)\right]\ge&\sup_{T\in\gT}\left\{\E_{s\sim\sP}[T(s)]-\E_{s\sim\sQ}[f^*(T(s))]\right\}
\end{aligned}
\end{equation*}
where $\gT\subseteq\{T:\gS\mapsto\sR\}$ is any class of functions and $f^*(z):=\sup_{u\in\sR}\{u\cdot z-f(u)\}$ for any $z\in\sR_+$ is the Fenchel conjugate.
\end{proposition}

\section{Sample Complexity of Optimization Problem~(\ref{obj-f})}

\begin{theorem}[Sample complexity of $f$-VI-IGL]
\label{thm_sc_f}
Consider a feedback-dependent reward decoder class $\Psi$ such that $\psi(y)\in[c,1-c]$ for any $\psi\in\Psi$ and $y\in\gY$, where $c\in(0,\frac{1}{2})$. Suppose the functions $|G|,|T|,|f_1^*(G)|,|f_2^*(T)|\le B^*\le\infty$ are bounded. Then, for any $\delta\in(0,1]$, given a dataset $\gD=\{(x_k,a_k,y_k)\}_{k=1}^K$ collected by the behavior policy $\pi_\textup{b}:\gX\to\Delta_\gA$, there exists an algorithm such that the solved reward decoder $\widehat{\psi}$ from the optimization problem~(\ref{opt_obj}) satisfies that $|\gL_f(\widehat{\psi})-\gL_f^*|$ is bounded by
\begin{equation*}
\begin{aligned}
    \max\{1,\beta\}\cdot O\left(\frac{(1-c)^2}{c^2}\cdot \sum_{y\in\gY^\epsilon_{\Xi}:\sigma_{\pi_\textup{b}}^\epsilon(y)>0}\sqrt{\frac{(B^*)^2|\gY^\epsilon_{\Xi}|^2}{K\cdot \sigma_{\pi_\textup{b}}^\epsilon(y)}\log\left(\frac{|\gY^\epsilon_{\Xi}|d_{\Psi,\gT,\gG}}{\delta}\right)}\right)
\end{aligned}
\end{equation*}
where $\gL_f^*$ is the optimal value to the optimization problem~(\ref{opt_obj}), parameters $\sigma_{\pi_\textup{b}}^\epsilon$ and $d_{\Psi,\gT,\gG}$ are defined in the Appendix~\ref{p_thm_sc}, and  $\gY^\epsilon_{\Xi}$ is $\epsilon$-covering of feedback space $\gY$ with respect to the joint function class $\Xi:=\Psi\cup\{f^*(G(x,a,\cdot,r)):(x,a,r)\in\gX\times\gA\times\{0,1\}\}_{G\in\gG}$.
\begin{proof}
    The proof follows the exact same analysis in Appendix~\ref{p_thm_sc}, with $f^*(x)=\exp(x-1)$ as a special case.
\end{proof}
\end{theorem}

\section{Additional Experimental Results}

\subsection{Robustness to Noises}
\label{sec:F.1}

This section provides the detailed data in Section~\ref{sec: emp_noises}. We compare the performance of \textsf{E2G}~\cite{pmlr-v139-xie21e} and the VI-IGL algorithm~\ref{alg0} in the number-guessing task.  We report both the policy accuracy and the standard deviation. The results are averaged over 16 trials. Specifically, Table~\ref{tab_0} corresponds to the noiseless setting and Tables~\ref{tab_1}$\sim$\ref{tab_1.2} show the results under three noise levels~($0.1,0.2,0.3$). We use the results for $\beta=10$ to plot Figure~\ref{fig:1}. 

In addition, the results show that as the noise level increases, our regularized objective~(\ref{obj}) with $\beta=10$ attains more consistent performance than the unregularized one~(i.e., only minimizing $I(X,A;Y|R_\psi)$, which is shown by $\beta=0$), in terms of both accuracy and the standard deviation. This reinforces the necessity to include the regularization term.

\begin{table}[!h]
\begin{center}
\begin{tabular}{ |l||c|c|c| }
\hline
\text{Methods}&
VI-IGL ($\beta=0$) & VI-IGL ($\beta=10$) & \textsf{E2G} \\
\hline
\hline
\texttt{N} & $55.7\pm27.3$ & $\bm{81.6\pm7.9}$ & $\bm{82.2\pm4.3}$ \\
\hline
\end{tabular}
\caption{Noiseless setting. The results are averaged over 16 trials.}
\label{tab_0}
\end{center}
\end{table}

\begin{table}[!h]
\begin{center}
\begin{tabular}{ |l||c|c|c| }
\hline
\text{Methods}&
VI-IGL ($\beta=0$) & VI-IGL ($\beta=10$) & \textsf{E2G} \\
\hline
\hline
\texttt{I} & $60.6\pm15.5$ &  $\bm{71.0\pm16.2}$   &  $25.0\pm18.4$   \\
\hline
\texttt{A} & $64.8\pm17.9$ &  $\bm{73.6\pm16.0}$   & $21.6\pm12.4$   \\
\hline
\texttt{C} & $52.4\pm25.3$ & $\bm{72.1\pm10.2}$   &  $15.4\pm14.3$  \\
\hline
\texttt{C-A} & $63.4\pm16.4$ & $\bm{65.2\pm16.3}$   & $20.6\pm14.8$  \\
\hline
\end{tabular}
\caption{Noise level=0.1. The results are averaged over 16 trials~(\texttt{I}: independent, \texttt{A}: action-inclusive, \texttt{C}: context-inclusive, \texttt{C-A}: context-action-inclusive).}
\label{tab_1}
\end{center}
\end{table}

\begin{table}[!h]
\begin{center}
\begin{tabular}{ |l||c|c|c| }
\hline
\text{Methods}&
VI-IGL ($\beta=0$) & VI-IGL ($\beta=10$)  & \textsf{E2G} \\
\hline
\hline
\texttt{I} & $54.6\pm23.5$  &  $\bm{68.8\pm16.2}$   &  $18.5\pm13.1$  \\
\hline
\texttt{A} & $43.9\pm23.5$ &  $\bm{64.1\pm18.7}$   &  $15.1\pm11.7$  \\
\hline
\texttt{C} & $49.0\pm25.9$  &  $\bm{62.7\pm21.6}$    &  $21.0\pm13.3$ \\
\hline
\texttt{C-A} & $57.6\pm24.3$ &  $\bm{62.8\pm23.9}$    & $17.2\pm13.7$ \\
\hline
\end{tabular}
\caption{Noise level=0.2. The results are averaged over 16 trials~(\texttt{I}: independent, \texttt{A}: action-inclusive, \texttt{C}: context-inclusive, \texttt{C-A}: context-action-inclusive).}
\label{tab_1.1}
\end{center}
\end{table}

\begin{table}[!h]
\begin{center}
\begin{tabular}{ |l||c|c|c| }
\hline
\text{Methods}&
VI-IGL ($\beta=0$) & VI-IGL ($\beta=10$) & \textsf{E2G} \\
\hline
\hline
\texttt{I} & $56.9\pm21.4$ & $\bm{57.8\pm19.0}$   &  $16.8\pm12.7$  \\
\hline
\texttt{A} & $33.6\pm22.3$ & $\bm{42.8\pm17.2}$  &  $16.4\pm12.6$  \\
\hline
\texttt{C} & $50.5\pm21.2$ & $\bm{51.4\pm18.3}$   &  $24.6\pm16.4$ \\
\hline
\texttt{C-A} & $\bm{50.6\pm22.6}$ & $\bm{50.3\pm18.1}$   & $18.5\pm14.7$ \\
\hline
\end{tabular}
\caption{Noise level=0.3. The results are averaged over 16 trials~(\texttt{I}: independent, \texttt{A}: action-inclusive, \texttt{C}: context-inclusive, \texttt{C-A}: context-action-inclusive).}
\label{tab_1.2}
\end{center}
\end{table}

\subsection{Value of Parameter $\beta$}
\label{sec:F.2}

This section provides the detailed data in Section~\ref{sec: 6.2}.

Tables~\ref{tab_E21} and \ref{tab_3} show the results for the noiseless setting and the noisy settings (with noise level 0.1), respectively. In contrast, the performance of the unregularized objective significantly degrades.

\begin{table}[!h]
\begin{center}
\begin{tabular}{ |l||c|c|c|c|c| }
\hline
$\beta$ & $0$ & $5$ & $10$ & $15$ & $20$ \\
\hline
\hline
\texttt{N} & $55.7\pm27.3$ & $69.7\pm15.6$  & $\bm{81.6\pm7.9}$& $79.1\pm9.1$ & $72.7\pm16.4$ \\
\hline
\end{tabular}
\caption{Value of Parameter $\beta$: Noiseless environment. The results are averaged over 16 trials.}
\label{tab_E21}
\end{center}
\end{table}

\begin{table}[!h]
\begin{center}
\begin{tabular}{ |l||c|c|c|c|c| }
\hline
$\beta$ & $0$ & $5$ & $10$ & $15$ & $20$ \\
\hline
\hline
\texttt{I} & $60.6\pm15.5$ & $59.5\pm24.2$ & $\bm{71.0\pm16.2}$ & $\bm{71.7\pm19.3}$ & $63.0\pm18.7$ \\
\hline
\texttt{A}  & $64.8\pm17.9$  & $68.5\pm18.8$ & $\bm{73.6\pm16.0}$& $67.7\pm20.6$  & $58.2\pm19.5$ \\
\hline
\texttt{C} &  $52.4\pm25.3$   & $61.4\pm13.9$ & $\bm{72.1\pm10.2}$& $63.1\pm21.1$  & $58.0\pm26.3$ \\
\hline
\texttt{C-A} & $63.4\pm16.4$ & $\bm{68.7\pm20.8}$ & $65.2\pm16.3$& $60.2\pm15.0$ & $62.6\pm11.7$ \\
\hline
\end{tabular}
\caption{Value of Parameter $\beta$: Noise level$=0.1$. The results are averaged over 16 trials~(\texttt{I}: independent, \texttt{A}: action-inclusive, \texttt{C}: context-inclusive, \texttt{C-A}: context-action-inclusive).}
\label{tab_3}
\end{center}
\end{table}

\begin{figure}[h]
\begin{center}
    \includegraphics[width=0.4\textwidth]{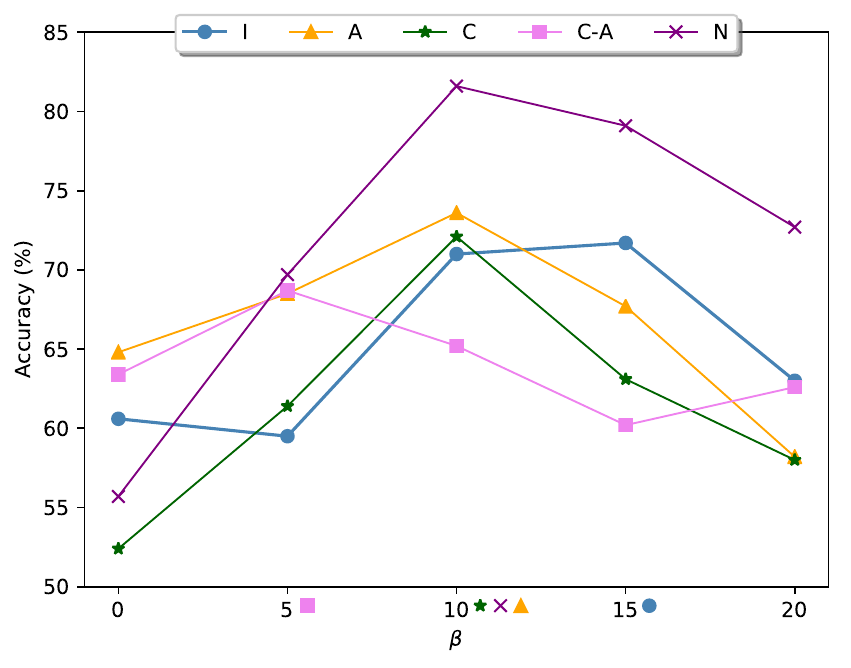}
\end{center}
\caption{Policy accuracy for different $\beta$: All noisy settings have level$=0.1$. The optimal selections are marked beside the value. The results show the necessity of regularization. The results are averaged over 16 trials.}
\label{fig:2}
\end{figure}

\section{Additional experimental details}
\label{App_A}

For the $f$-variational estimators~(functions $T$ and $G$), the reward decoder $\psi$, and the linear policy $\pi$, we use a 2-layer fully-connected network to process each input image~(i.e., the context or the feedback). Then, the concatenated inputs go through an additional linear layer and the final value is output. The same network structures are used to implement the reward decoder and the policy of the previous IGL algorithm~\citep{pmlr-v139-xie21e}. In each experiment, we train the $f$-VI-IGL algorithm for $1,000$ epochs with a batch size of $600$. Particularly, we alternatively update the parameters of the $f$-MI estimators and the reward decoders~(i.e., $500$ epochs of training for each). To stabilize the training, we clip the gradient norm to be no greater than $1$ and use an exponential moving average~(EMA) with a rate of $0.99$. For the previous IGL method, we follow the experimental details provided in the work of Xie~et~al.~\citep[Appendix C]{pmlr-v139-xie21e} and train the algorithm for 10 epochs over the entire training datasets.

%% file: main.bbl
\begin{thebibliography}{}

\bibitem[Ali and Silvey, 1966]{f3b5d8df-184a-3263-b0e6-61806ef005a0}
Ali, S.~M. and Silvey, S.~D. (1966).
\newblock A general class of coefficients of divergence of one distribution from another.
\newblock {\em Journal of the Royal Statistical Society. Series B (Methodological)}, 28(1):131--142.

\bibitem[Belghazi et~al., 2018]{pmlr-v80-belghazi18a}
Belghazi, M.~I., Baratin, A., Rajeshwar, S., Ozair, S., Bengio, Y., Courville, A., and Hjelm, D. (2018).
\newblock Mutual information neural estimation.
\newblock In Dy, J. and Krause, A., editors, {\em Proceedings of the 35th International Conference on Machine Learning}, volume~80 of {\em Proceedings of Machine Learning Research}, pages 531--540. PMLR.

\bibitem[Cohen et~al., 2017]{cohen2017emnist}
Cohen, G., Afshar, S., Tapson, J., and van Schaik, A. (2017).
\newblock Emnist: an extension of mnist to handwritten letters.

\bibitem[Csisz{\'a}r, 1967]{csiszar2022information}
Csisz{\'a}r, I. (1967).
\newblock Information-type measures of difference of probability distributions and indirect observation.
\newblock {\em studia scientiarum Mathematicarum Hungarica}, 2:229--318.

\bibitem[Donsker and Varadhan, 1983]{c52c83e0b02c4746a5ea29b5cd44fd00}
Donsker, M. and Varadhan, S. (1983).
\newblock Asymptotic evaluation of certain markov process expectations for large time. iv.
\newblock {\em Communications on Pure and Applied Mathematics}, 36(2):183--212.

\bibitem[Dud{\'{\i}}k et~al., 2014]{Dud_k_2014}
Dud{\'{\i}}k, M., Erhan, D., Langford, J., and Li, L. (2014).
\newblock Doubly robust policy evaluation and optimization.
\newblock {\em Statistical Science}, 29(4).

\bibitem[Dudik et~al., 2011]{dudik2011efficient}
Dudik, M., Hsu, D., Kale, S., Karampatziakis, N., Langford, J., Reyzin, L., and Zhang, T. (2011).
\newblock Efficient optimal learning for contextual bandits.

\bibitem[Gao et~al., 2015]{pmlr-v38-gao15}
Gao, S., Ver~Steeg, G., and Galstyan, A. (2015).
\newblock {Efficient Estimation of Mutual Information for Strongly Dependent Variables}.
\newblock In Lebanon, G. and Vishwanathan, S. V.~N., editors, {\em Proceedings of the Eighteenth International Conference on Artificial Intelligence and Statistics}, volume~38 of {\em Proceedings of Machine Learning Research}, pages 277--286, San Diego, California, USA. PMLR.

\bibitem[Gregor et~al., 2016]{gregor2016variational}
Gregor, K., Rezende, D.~J., and Wierstra, D. (2016).
\newblock Variational intrinsic control.

\bibitem[Kullback, 1997]{kullback1997information}
Kullback, S. (1997).
\newblock {\em Information theory and statistics}.
\newblock Courier Corporation.

\bibitem[Langford and Zhang, 2007]{NIPS2007_4b04a686}
Langford, J. and Zhang, T. (2007).
\newblock The epoch-greedy algorithm for multi-armed bandits with side information.
\newblock In Platt, J., Koller, D., Singer, Y., and Roweis, S., editors, {\em Advances in Neural Information Processing Systems}, volume~20. Curran Associates, Inc.

\bibitem[Lecun et~al., 1998]{Lecun1998}
Lecun, Y., Bottou, L., Bengio, Y., and Haffner, P. (1998).
\newblock Gradient-based learning applied to document recognition.
\newblock {\em Proceedings of the IEEE}, 86(11):2278--2324.

\bibitem[Levine et~al., 2011]{NIPS2011_c51ce410}
Levine, S., Popovic, Z., and Koltun, V. (2011).
\newblock Nonlinear inverse reinforcement learning with gaussian processes.
\newblock In Shawe-Taylor, J., Zemel, R., Bartlett, P., Pereira, F., and Weinberger, K., editors, {\em Advances in Neural Information Processing Systems}, volume~24. Curran Associates, Inc.

\bibitem[Maghakian et~al., 2023]{maghakian2023personalized}
Maghakian, J., Mineiro, P., Panaganti, K., Rucker, M., Saran, A., and Tan, C. (2023).
\newblock Personalized reward learning with interaction-grounded learning ({IGL}).
\newblock In {\em The Eleventh International Conference on Learning Representations}.

\bibitem[Molavipour et~al., 2020]{9053422}
Molavipour, S., Bassi, G., and Skoglund, M. (2020).
\newblock Conditional mutual information neural estimator.
\newblock In {\em ICASSP 2020 - 2020 IEEE International Conference on Acoustics, Speech and Signal Processing (ICASSP)}, pages 5025--5029.

\bibitem[Ng and Russell, 2000]{10.5555/645529.657801}
Ng, A.~Y. and Russell, S.~J. (2000).
\newblock Algorithms for inverse reinforcement learning.
\newblock In {\em Proceedings of the Seventeenth International Conference on Machine Learning}, ICML '00, page 663–670, San Francisco, CA, USA. Morgan Kaufmann Publishers Inc.

\bibitem[Nguyen et~al., 2010]{10.1109/TIT.2010.2068870}
Nguyen, X., Wainwright, M.~J., and Jordan, M.~I. (2010).
\newblock Estimating divergence functionals and the likelihood ratio by convex risk minimization.
\newblock {\em IEEE Trans. Inf. Theor.}, 56(11):5847–5861.

\bibitem[Paninski, 2003]{10.1162/089976603321780272}
Paninski, L. (2003).
\newblock Estimation of entropy and mutual information.
\newblock {\em Neural Comput.}, 15(6):1191–1253.

\bibitem[Peason, 1900]{doi:10.1080/14786440009463897}
Peason, K. (1900).
\newblock On the criterion that a given system of deviations from the probable in the case of a correlated system of variables is such that it can be reasonably supposed to have arisen from random sampling.
\newblock {\em The London, Edinburgh, and Dublin Philosophical Magazine and Journal of Science}, 50(302):157--175.

\bibitem[Russo and Van~Roy, 2014]{NIPS2014_301ad0e3}
Russo, D. and Van~Roy, B. (2014).
\newblock Learning to optimize via information-directed sampling.
\newblock In Ghahramani, Z., Welling, M., Cortes, C., Lawrence, N., and Weinberger, K., editors, {\em Advances in Neural Information Processing Systems}, volume~27. Curran Associates, Inc.

\bibitem[Schalk et~al., 2004]{1300799}
Schalk, G., McFarland, D., Hinterberger, T., Birbaumer, N., and Wolpaw, J. (2004).
\newblock Bci2000: a general-purpose brain-computer interface (bci) system.
\newblock {\em IEEE Transactions on Biomedical Engineering}, 51(6):1034--1043.

\bibitem[Serrhini and Dargham, 2017]{10.1007/978-3-319-46568-5_14}
Serrhini, M. and Dargham, A. (2017).
\newblock Toward incorporating bio-signals in online education case of assessing student attention with bci.
\newblock In Rocha, {\'A}., Serrhini, M., and Felgueiras, C., editors, {\em Europe and MENA Cooperation Advances in Information and Communication Technologies}, pages 135--146, Cham. Springer International Publishing.

\bibitem[Sutton and Barto, 2018]{sutton2018reinforcement}
Sutton, R.~S. and Barto, A.~G. (2018).
\newblock {\em Reinforcement learning: An introduction}.
\newblock MIT press.

\bibitem[Tishby et~al., 2000]{tishby2000information}
Tishby, N., Pereira, F.~C., and Bialek, W. (2000).
\newblock The information bottleneck method.

\bibitem[Xie et~al., 2021a]{xie2021policy}
Xie, T., Jiang, N., Wang, H., Xiong, C., and Bai, Y. (2021a).
\newblock Policy finetuning: Bridging sample-efficient offline and online reinforcement learning.
\newblock In Beygelzimer, A., Dauphin, Y., Liang, P., and Vaughan, J.~W., editors, {\em Advances in Neural Information Processing Systems}.

\bibitem[Xie et~al., 2021b]{pmlr-v139-xie21e}
Xie, T., Langford, J., Mineiro, P., and Momennejad, I. (2021b).
\newblock Interaction-grounded learning.
\newblock In Meila, M. and Zhang, T., editors, {\em Proceedings of the 38th International Conference on Machine Learning}, volume 139 of {\em Proceedings of Machine Learning Research}, pages 11414--11423. PMLR.

\bibitem[Xie et~al., 2022]{xie2022interactiongrounded}
Xie, T., Saran, A., Foster, D.~J., Molu, L.~P., Momennejad, I., Jiang, N., Mineiro, P., and Langford, J. (2022).
\newblock Interaction-grounded learning with action-inclusive feedback.
\newblock In Oh, A.~H., Agarwal, A., Belgrave, D., and Cho, K., editors, {\em Advances in Neural Information Processing Systems}.

\end{thebibliography}
